\documentclass[review]{elsarticle}

\usepackage{hyperref}
\usepackage{natbib}
\setcitestyle{square,sort,comma,numbers}

\usepackage{placeins}
\usepackage{array}
\usepackage[utf8]{inputenc}  
\usepackage[T1]{fontenc}  
\usepackage{amsmath}
\usepackage{amssymb}

\usepackage{tabularx}
\usepackage{multirow}
\usepackage{graphicx}
\usepackage{subfigure}
\usepackage{epstopdf}
\usepackage{diagbox}
\usepackage{lscape}
\usepackage[english]{babel}
\usepackage{bbm}
\usepackage{algorithmic}
\usepackage{algorithm}

\usepackage{tabularx}
\usepackage{floatflt,amssymb}
\usepackage{multirow} %% pour regrouper un texte sur plusieurs lignes dans une table
\usepackage{url} %% pour citer les url par \url
\usepackage{tikz}
    \usetikzlibrary{shapes.arrows}
    \usetikzlibrary{decorations.shapes}
    \usetikzlibrary{decorations.pathreplacing}
    \usetikzlibrary{fadings,shapes.arrows,shadows}
\makeatletter
\DeclareRobustCommand\sfrac[1]{\@ifnextchar/{\@sfrac{#1}}%
                                            {\@sfrac{#1}/}}
\def\@sfrac#1/#2{\leavevmode\kern.1em\raise.5ex
         \hbox{$\m@th\mbox{\fontsize\sf@size\z@
                           \selectfont#1}$}\kern-.1em
         /\kern-.15em\lower.25ex
          \hbox{$\m@th\mbox{\fontsize\sf@size\z@
                            \selectfont#2}$}}

\usetikzlibrary{positioning}

\makeatletter
\renewcommand*\env@matrix[1][*\c@MaxMatrixCols c]{%
  \hskip -\arraycolsep
  \let\@ifnextchar\new@ifnextchar
  \array{#1}}
\makeatother

\usepackage{xcolor}
\definecolor{newcolor}{rgb}{.8,.349,.1}
\definecolor{mygreen}{rgb}{0.2, 0.65, 0.15}

\newcommand{\x}{\mathbf{x}}
\newcommand{\z}{\mathbf{z}}
\newcommand{\ba}{\mathbf{a}}
\newcommand{\D}{\mathbf{D}}
\newcommand{\M}{\mathbf{M}}
\newcommand{\A}{\mathbf{A}}

\newcommand{\corGG}[2]{{\color{black}#2}}
\newcommand{\removeGG}[1]{}

\usepackage{caption}
\DeclareCaptionFont{white}{\color{white}}
\DeclareCaptionFormat{listing}{\colorbox[cmyk]{0.43, 0.35, 0.35,0.01}{\parbox{\textwidth}{\hspace{15pt}#1#2#3}}}
\usepackage[k-loose]{minitoc}

\usepackage{textcomp}

\usepackage{amsthm}

\usepackage{lipsum} % Used for inserting dummy 'Lorem ipsum' text into the template
\newcommand{\norme}[1]{\left\Vert #1\right\Vert}

\newcommand{\argmin}{\operatornamewithlimits{argmin}}

\usepackage[normalem]{ulem}

\usepackage[toc,page]{appendix}

\usepackage{booktabs}

\usepackage{enumitem}
\def\un{ {\mathrm{I\hskip-5.9pt 1}}}

\journal{}

\biboptions{authoryear}

\begin{document}

\begin{frontmatter}

\title{An efficient supervised dictionary learning method for audio signal recognition}

\author[]{Imad Rida\corref{mycorrespondingauthor}}
\ead{imad.rida@insa-rouen.fr}
\author[]{Romain H\'{e}rault}
\ead{romain.herault@insa-rouen.fr}
\author[]{Gilles Gasso}
\ead{gilles.gasso@insa-rouen.fr}
\cortext[mycorrespondingauthor]{Corresponding author}

\address{Normandie Universit\'{e}, LITIS EA 4108, INSA Rouen Normandie, Avenue de l'Universit\'{e}, 76800, Saint Etienne du Rouvray, France}

\begin{abstract}
Machine hearing or listening represents an emerging area.
Conventional approaches rely on the design of handcrafted features specialized to a specific audio task and that can hardly generalized to other audio fields.
For example, Mel-Frequency Cepstral Coefficients (MFCCs) and its variants were successfully applied to computational auditory scene recognition while Chroma vectors  are good at music chord recognition. 
Unfortunately, these predefined features may be of variable discrimination power while extended to other tasks or even within the same task due to different nature of clips.
Motivated by\removeGG{this fact and due to the} this need \corGG{in practice to a machine hearing framework operating in various application domains}{of a principled framework across domain applications for machine listening}, we propose a generic and data-driven representation learning approach.
For this sake, a novel and efficient supervised dictionary learning method is presented.
The \removeGG{proposed} method \corGG{attempts to learn}{learns} dissimilar \corGG{dictionary}{dictionaries}, one per each class, in order to extract heterogeneous information for classification.
In other words, we are seeking to minimize the intra-class homogeneity and maximize class separability.
This is made possible by promoting pairwise orthogonality between class specific dictionaries and controlling the \corGG{sparsity coefficients structure}{sparsity structure of the audio clip's decomposition over these dictionaries}.
The resulting optimization problem is non-convex and solved using a proximal gradient descent method.
Experiments are performed on both computational auditory scene (East Anglia and Rouen) and synthetic music chord recognition datasets.
Obtained results show that our method is capable to reach state-of-the-art hand-crafted features for both applications.
\end{abstract}


\end{frontmatter}

%\linenumbers

\section{Introduction} \label{intro} 
Humans have a very high perception capability through physical sensation, which can include sensory input from the eyes, ears, nose, tongue, or skin. A lot of efforts have been devoted to develop intelligent computer systems capable to interpret data in a similar manner to the way humans use their senses to relate to the world around them \cite{rida2015unsupervised}. While most efforts have focused on vision perception which \corGG{represents}{is} the dominant sense in humans, machine listening \corGG{}{(ability of a  machine to fully understand an audio input)} represents an emerging area \cite{lyon2010machine}. One rising application domain we are interested in is the classification of environmental audio signals usually termed as Computational Auditory Scene Recognition (CASR). It refers to the task of associating a semantic label to an audio stream that identifies the environment in which it has been produced. Another application is music recognition especially chords recognition that represent the most fundamental structure and the back-bone of occidental music. 

The usual trend to classify signals is first to extract discriminative feature representations from the signals, and then feed a classifier with them \cite{rida2018palmprintt,rida2015improved,rida2018ensemble}. In this case, features are chosen so as to enforce similarities within a class and disparities between classes \cite{rida2016gaitt}. The more discriminative the features are, the better the classifier performs. Because of the specific peculiarities of audio clips in different application domains, specialized features have to be designed. For instance, chroma vectors represent the dominant representation in order to extract the harmonic contents from music signals \cite{fujichroma,sheh2003chord, oudre2011chord,sparsechroma,Kronvall2017105}. In audio scene recognition, recorded signals can be potentially composed of a very large amount of sound events. To tackle this problem, features such as Mel-Frequency Cepstral Coefficients (MFCCs) and its variants \cite{davis1980comparison, kinnunen2012low,zheng2001comparison,Benzeghiba2007763,li2014overview,rida2018feature} have been successfully combined with different classification techniques. These predefined features may be of variable discrimination power while extended to other tasks or different nature of clips. For this reason and due to the need to a machine hearing framework operating in various application domains, the suited feature representations should be automatically learned. 

In recent years there has been a growing interest in the study of sparse representation learning. Using an overcomplete dictionary that contains prototype signal-atoms, signals are described as linear combinations of a few of these atoms. Audio representation learning techniques can be broadly divided into four main approaches \cite{sangnier2015filter}: wavelets \cite{strauss2003feature,yger2011wavelet}, Cohen distribution \cite{davy2002optimized,honeine2006optimal}, dictionary  \cite{mairal2009,ramirez2010} and filter banks \cite{biem2001application,sangnier2015filter}. Choosing a pre-specified transform matrix is appealing because it is simpler. Also, in many cases it leads to simple and fast algorithms for the evaluation of the sparse representation. This is indeed the case for overcomplete wavelets, Cohen and filter banks. The success of such dictionaries in applications depends on how suitable they are to sparsely describe the signals in question.

%\begin{table}[h]
  %\centering
  %\caption {Non exhaustive sparse representation learning \cite{sangnier2015filter}.}
%\resizebox{0.45\linewidth}{!}{
 % \begin{tabular}{llll}
   % \multicolumn{2}{c}{}  \\
      %  \toprule
       %  \toprule
   % Approach & Methods   \\
   % \midrule  
    % \midrule  
      %    \tabitem  Wavelets                           & \tabitem  \cite{strauss2003feature,yger2011wavelet}\\
     
         %                          \hline
            %                       \hline
  %  \tabitem Cohen Distribution  & \tabitem \cite{davy2002optimized,honeine2006optimal} \\
     %                                \hline
        %                             \hline
    % \tabitem Dictionary   & \tabitem \cite{mairal2009,ramirez2010} \\
       %                               \hline
          %                            \hline
   %  \tabitem Filter Bank     &  \tabitem \cite{biem2001application,sangnier2015filter} \\                                         
   % \bottomrule
   %  \bottomrule
 % \end{tabular}}
  %\label{tab:summtff}
%\end{table}

Recently, a different route for designing dictionaries based on learning is considered. It seeks to find the dictionary $\mathbf{D}$ that yields sparse representations for the training signals. Such dictionaries have the potential to outperform commonly used pre-determined dictionaries \cite{aharon2006img}.

\section{Motivations and Contributions}
%Sparse representation of signals and images has known a big interest from researchers in order to analyze, extract or select features.  A "sparse representation" means that a signal or image can be represented as a linear combination of few representative elements, called dictionary atoms. The main challenge of the sparse representation is the choice of the dictionary on which the signal will be represented and the sparsity type. The simplest approach to tackle this problem is to take predefined dictionary such as wavelet analysis, Gabor atoms or Discrete Cosine Basis, but this will give us no guarantee that these predefined dictionaries will be able to represent and extract useful information for the problem in question.  One possible solution to tackle this problem is to learn the suited set of dictionary atoms from the data. 

Conventional dictionary learning formulation minimizes the reconstruction error between a given signal and its (sparse) representation over the learned dictionary.  Although this formulation is convenient for solving signal denoising \cite{elad2006,mairal2008,eksioglu2014online}, inpainting \cite{elad2010} and segmentation \cite{flores2016segmentation} problems, it may not suit classification tasks where the ultimate goal is to get discriminative decomposition of training signals over the learned dictionary \cite{phaisangittisagul2017predictive,al2018palmprint,rida2018comprehensive}. Motivated by the limitation of the conventional dictionary learning techniques for classification, supervised dictionary learning has known a wide emergence. Related  techniques can be organized in six main groups \cite{gangeh2015supervised} summarized in Table \ref{tab:summtf}.

\begin{table}[ht]
	\centering
	\caption {Summary of supervised dictionary learning techniques for data classification \cite{gangeh2015supervised}.}
	%  \noindent
	\resizebox{0.95\linewidth}{!}{
		\begin{tabular}{llll}
			%  \toprule
			\multicolumn{3}{c}{}  \\
			\toprule
			\toprule
			Ref & Approach  & Advantages  \& Limitations  \\
			\midrule  
			\midrule

			\cite{yang2010}               &  \textbf{A.}~  Dictionary per class   &  $(+)$ ease dictionary computation   \\
			\cite {ramirez2010}        &                                                      &    $(-)$     very large dictionary \\
			\hline
			\hline
			
			\cite{fulkerson2008localizing} &  \textbf{B.}~   Prune large dictionaries     & $(+)$ ease dictionary computation \\
			\cite{winn2005object}                &                                                               & $(-)$ low performances  \\
			\hline
			\hline
			
			\cite{mairal2009}     &   \textbf{C.}~  Joint dictionary \& classifier learning  &  $(+)$ good performances \\
			\cite{zhang2010}     &                                                                                  & $(-)$ too many parameters     \\
			\hline
			\hline
			
			\cite{zhang2013simultaneous}  &   \textbf{D.}~   Labels in dictionary   & $(+)$ good performances \\
			\cite{lazebnik2009supervised}   &                                                         &  $(-)$ complex optimization  \\
			
			\hline
			\hline
			
			\cite{yang2011}     &   \textbf{E.}~     Labels in coefficients  &  $(+)$ good performances     \\
			&                              & $(-)$ complex \\
			
			\hline
			\hline

			\cite {varma2009statistical}                                                                               &    \textbf{F.}~  Histograms of dictionary elements   & $(+)$ good performances \\                                                                                                                                                                                                                                         
			\cite{lian2010probabilistic}           &   & $(-)$ only based local constituents   \\
			
			\bottomrule
			\bottomrule
	\end{tabular}}
	\label{tab:summtf}
\end{table}

\begin{itemize}

\item Learning one dictionary per class

Seeks to learn a dictionary per class \cite{yang2010,tuysuzouglu2018sparse}. Although this approach can be potentially performing, learned dictionaries can capture similar properties for different classes leading to poor classification performance. To tackle this problem, \cite {ramirez2010} suggested to make the learned dictionaries as different as possible by enforcing their orthoganility to capture distinct information. A new test sample is assigned to class label of the dictionary providing the minimal residual reconstruction error.

\item Prune large dictionaries

In this approach, a very large dictionary is learned, then the dictionary atoms are merged based on a predefined criterion including Agglomerative Information Bottleneck (AIB) \cite{fulkerson2008localizing} and Mutual Information (MI) \cite{winn2005object}.

\item Joint dictionary and classifier learning

This approach seeks to jointly learn the classifier parameters and dictionary \cite{mairal2009,zhang2010}.

\item Embedding class labels into the learning of dictionary

In this approach, the data is first projected into a space where the intra and inter-class are minimized and maximized respectively, and subsequently learn the dictionary and the sparse representation in this new space \cite{zhang2013simultaneous,lazebnik2009supervised}.

\item Embedding class labels into the learning of sparse coefficients

This approach seeks to include class labels in the learning of coefficients. It is based on the minimization and the maximization of the within-class and the between-class covariance of the coefficients respectively \cite{yang2011}.

\item Learning a histogram of dictionary elements over signal constituents

In this approach a histogram of dictionary atoms learned on local constituents is computed. The resulting histograms are used to train a classifier and predict the class label of a new test signal \cite {varma2009statistical} \cite{lian2010probabilistic}. 
\end{itemize}

%\bigskip

Based on \corGG{the brief study of supervised dictionary approaches}{the characteristics of these methods,}, we introduce in the following a novel supervised dictionary method. Our proposed \corGG{method tries to exploit the strong points }{approach aims to exploit the strengths} of the previous methods that is: i) learning one dictionary per class, and ii) embedding class labels to force \corGG{sparse coefficients}{sparsity pattern of the signal's representation}. To this end, we encourage the dissimilarity between the dictionaries by penalizing the pairwise similarity between them. To reach superior discrimination power, we push towards zero the coefficients of a signal representation over other dictionaries than the one corresponding to its class label. \corGG{}{The contributions of the paper are: 
\begin{itemize}
		\item a novel supervised dictionary learning formulation,
		\item a related optimization algorithm based on alternating a sparse coding step with the update step of the dictionaries,
		\item experimental evaluations on scene and chord recognition applications.
\end{itemize}}

%\begin{landscape}

The remainder of this paper is organized as follows. Section \ref{sec:ourcontribution} \corGG{introduces the theoretical description of}{describes} the proposed method. Section \ref{exp} reports the experimental results and discussions. Finally, Section \ref{conc} concludes the paper.

\section{Proposed approach}
\label{sec:ourcontribution}
Let consider $\{(\mathbf{x}_{n},y_{n})\}_{n=1}^{N}$ where $\mathbf{x}_{n} \in \mathbb{R}^{M}$ is a signal and $y_{n} \in \{1,\cdots, C\}$ its label. \corGG{}{Our novel approach for supervised dictionary learning seeks to learn $C$ incoherent dictionaries $\D_c$, each per class, by enforcing their pairwise orthogonality. Furthermore to render the representation of a signal $\x$ with label $y=c'$ specific to its class, the coefficients of its decomposition over dictionaries $\D_c, c \neq c'$ are pushed towards zero. To illustrate the intuition behind the approach, let suppose a binary classification problem. Given a sample $(\x, y=1)$, we aim to find a decomposition  $\mathbf{x}   \approx \mathbf{\textcolor{black}{D}}_{\textcolor{black}{1}} \mathbf{\textcolor{black}{a}}_{\textcolor{black}{1}} +  \mathbf{\textcolor{black}{D}}_{\textcolor{black}{2}} \mathbf{\textcolor{black}{a}}_{\textcolor{black}{2}} $ such that the term $\norme{\mathbf{\textcolor{black}{D}}_{1}^{T} \mathbf{\textcolor{black}{D}}_{2}}_{F}^{2}$ reflecting the coherence between the dictionaries is small while enforcing the representation over $\D_2$ to be negligible by pushing the term  $\| \mathbf{\textcolor{black}{a}}_{\textcolor{black}{2}}\|_2^2$ close to zero. The obtained representations of the signals are further used as features in a linear SVM \cite{scholkopf2002learning}.

Before delving into the detailed formulation of the proposed approach and the way the involved optimization problem is addressed, we introduce the conventional dictionary learning method and its limitations. The following notations will be adopted: $\|\z \|_p = \sum_{j} \sqrt[1/p]{|z_j|^p} , p \geq 1$ stands for the $\ell_p$-norm of vector $\z$ and $\|\M\|_F = \sqrt[1/2]{\sum_{i,j} M_{ij}^2}$ represents the Frobenius norm of matrix $\M$. Finally the indicator function $\, \un_{y = c}$ is 1 if the inner condition is true, and 0 otherwise.
}

\subsection {Conventional dictionary learning}
\corGG{}{Dictionary learning was primary devised to find a linear decomposition of a signal using a few atoms of a learned overcomplete dictionary \cite{elad2006}. }
Let suppose a dictionary $\mathbf{D} \in \mathbb {R}^{M \times K}$ composed of $K$ atoms $\{\mathbf{d}_{k} \in \mathbb{R}^{M}\}_{k=1}^{K} $. The conventional approach seeks a sparse representation $\mathbf{a}_{n} \in \mathbb{R}^{K}$ of a signal $\mathbf{x}_{n} \in \mathbb {R}^{M}$ over $\mathbf{D}$ \corGG{such as:\\
	
	\begin{equation}
	\mathbf{x}_{n} \approx \sum_{k=1}^{K} a_{nk} \mathbf{d}_{k} = \mathbf{D} \mathbf{a}_{n}
	\end{equation}}{such as $\mathbf{x}_{n} \approx \sum_{k=1}^{K} a_{nk} \mathbf{d}_{k} \approx \mathbf{D} \mathbf{a}_{n}$. }
 Given a set of $N$ signals $\{\mathbf{x}_{n}\}_{n=1}^{N}$, \corGG{the coefficients of $\mathbf{a}_{n}$ as well as the dictionary $\mathbf {D}$ are obtained}{dictionary learning method intends to find simultaneously the dictionary $\mathbf {D}$ and the sparse codes $\mathbf{a}_{n}$} by solving the following optimization problem
 
 \begin{equation}
 \left\{
    \begin{array}{ll}
       \displaystyle \min_{\mathbf{D}, \{\mathbf{a}_{n}\}_{n=1}^{N}} \sum_{n=1}^{N} \norme{\mathbf{x}_{n}-\mathbf{D} \mathbf{a}_{n}}_{2}^{2} + \lambda \norme{\mathbf{a}_{n}}_{1}\\
       & \\
   \displaystyle  \hspace{0.4cm} \mbox{s.t} \hspace {0.5 cm} \norme{\mathbf{d}_{k}}_{2}^{2} \leq 1  \hspace {0.5 cm} \forall k=1, \cdots, K 
   \end{array}
\right.
\label{eq:conventional}
\end{equation} 
Formulation (\ref{eq:conventional}) is not suitable for classification since it solely seeks to minimize the reconstruction error between the input signal and its \corGG{sparse representation over the learned dictionary}{representation over the dictionary} \cite{7350221,8244323,rida2018novel}. In the following we extend this formulation \corGG{to tackle the various classification problems.}{to take into account the label information. Instead of determining a single global dictionary we focus in learning class specific dictionaries as presented in the next subsection.}

\subsection{Formulation of the supervised dictionary learning problem}

We consider a dictionary $\mathbf {D}_{c} \in \mathbb {R}^{M \times K'}$ associated to each class $c$. The global dictionary $\mathbf{D}= [ \mathbf{D}_{1} \cdots \mathbf{D}_{C} ] \in \mathbb{R}^{M \times K}$ represents the concatenation of the class based dictionaries $\{\mathbf{D}_{c}\}_{c=1}^{C}$. Each dictionary $\mathbf{D}_{c}$ is composed of $K'$ atoms $\{\mathbf{d}_{k} \in \mathbb{R}^{M}\}_{k=1}^{K'} $. \corGG{For simplicity sake}{For simplicity sake and without loss of generality} we consider $K'$ is the same for all $\{\mathbf{D}_{c}\}_{c=1}^{C}$. \corGG{The sparse representation of $\mathbf{x}_{n}$ over the global dictionary $\mathbf{D}$ is $\mathbf{a}_{n}^{T}= [ \mathbf{a}_{n1}^{T} \cdots \mathbf{a}_{nc}^{T} \cdots \mathbf{a}_{nC}^{T}]$ where $\mathbf{a}_{nc}$ represents the sparse representation over the class specific dictionary $\mathbf{D}_{c}$}{We assume the decomposition of $\x_n$ over the global dictionary $\mathbf{D}$ is given by $\x_n \approx \D \ba_n \approx \sum_{c=1}^n \D_c \, \ba_{nc}$ where the vector $\mathbf{a}_{n}^{T}= [ \mathbf{a}_{n1}^{T} \cdots \mathbf{a}_{nc}^{T} \cdots \mathbf{a}_{nC}^{T}]$ represents the overall sparse code of $\x_n$ and $\mathbf{a}_{nc} \in \mathbb{R}^{K'}$ represents its sparse representation over the class specific dictionary $\mathbf{D}_{c}$}. \corGG{Hence the sparse representation of the overall training data $\{\mathbf{x}_{n}\}_{n=1}^{N}$ is gathered in $\mathbf{A}=[ \mathbf{a}_{1} \cdots \mathbf{a}_{n}]$. The supervised dictionary learning problem we intend to address is formulated as follows:}{The supervised dictionary learning problem we intend to address seeks to: 
	\begin{itemize}
		\item capture as much as possible information in the signal by minimizing the global reconstruction error over $\D$;
		\item specialize the extracted information per class by minimizing the class specific reconstruction error similar to the minimization of intra-class homogeneity;
		\item render dissimilar the extracted class specific information by promoting pairwise orthogonality between dictionaries and "zeroing" coefficients not specific to the signal label. In other words, we attempt to maximize class separability; and
		\item promote the sparsity of signal representations over the dictionaries to preserve generalization ability of the linear SVM built upon  the sparse codes.
\end{itemize}
Let assume the coefficients related to the training signals $\{\mathbf{x}_{n}\}_{n=1}^{N}$ are gathered in $\mathbf{A}=[ \mathbf{a}_{1} \cdots \mathbf{a}_{n}]$. The dictionaries $\{\mathbf{D}_{c}\}_{c=1}^{C}$ and the codes $\A$ are obtained by solving the optimization problem 
}

\begin{equation}
 \left\{
    \begin{array}{ll}
      \displaystyle  \min_{\{\mathbf{D}_{c}\}_{c=1}^{C}, \{\mathbf{a}_{n}\}_{n=1}^{N}} J(\mathbf{A}, \D) = J_{1}(\D, \A) + \mu J_{2}(\D, \A) + \lambda J_{3}(\A) +\gamma_{1} J_{4}(\A)+ \gamma_{2} J_{5}(\D)\\
       & \\
   \displaystyle  s.t \hspace {0.5 cm} \norme{\mathbf{d}_{ck}}_{2}^{2} \leq 1  \hspace {0.5 cm} \forall c=1, \cdots, C \hspace {0.5 cm} \mbox{and} \hspace {0.5 cm} \forall k=1, \cdots, K \\ \\
   \end{array}
\right.
\label{eq:2}
\end{equation}
\corGG{where in the problem (\ref{eq:2})}{The terms included in problem (\ref{eq:2}) are defined as follows: }
 $$J_{1}(\D, \A)=\displaystyle \sum_{n=1}^{N} \norme{\mathbf{x}_{n}-\mathbf{D} \mathbf{a}_{n}}_{2}^{2}$$
 \noindent \corGG{represents}{measures} the global reconstruction error \corGG{}{of all training signals} over the global dictionary $\mathbf{D}$. \corGG{}{It is intended to capture the common patterns of the signals shared across different classes. The term}
$$J_{2}(\D, \A)=\displaystyle \sum_{n=1}^{N}  \sum_{c=1}^{C} \un_{y_{n}=c} \norme{\mathbf{x}_{n}-\mathbf{D}_{c} \mathbf{a}_{nc}}_{2}^{2}$$
\noindent stands for the class specific reconstruction error over the dictionary $\mathbf{D}_{c}$. In other words $J_{2}$ measures the quality of reconstructing a sample $(\mathbf{x}_{n}, \mathbf{y}_{n}=c)$ over the sole dictionary $\mathbf{D}_{c}$. \corGG{}{It aims to minimize intra-class homogeneity.
	
 Beyond these fitting errors, our learning scheme involves some regularization terms. The first one}
 
$$J_{3}(\A)=\displaystyle  \sum_{n=1}^{N}  \norme{\mathbf{a}_{n}}_{1}$$
 \noindent  is the classical sparsity \corGG{penalization}{regularization in overcomplete dictionary learning while}
 
$$J_{4}(\A)=\displaystyle \sum_{n=1}^{N} \sum_{c=1}^{C}  \un_{y_{n} \neq c} \norme{\mathbf {a}_{nc}}_{2}^{2}$$
\noindent  aims to push towards zero the coefficients  $\mathbf{a}_{nc}$ of the signal $\mathbf{x}_{n}$ representation over non-class specific dictionary $\mathbf{D}_{j}$, $j \neq y_{n}$. Finally

$$J_{5}(\D)=\displaystyle \sum_{c=1}^{C} \sum_{\substack {c'=1 \\ c' \neq c}}^{C} \norme {\mathbf{D}_{c}^{T} \mathbf{D}_{c'}}_{F}^{2}$$
 \noindent \corGG{with $\norme{.}_{F}$ is the Frobenius norm,}{} encourages the pairwise orthogonality between different dictionaries. \corGG{}{The last two regularization terms are deemed to promote large class separability of the learned coefficients.}
 
%To sum up, our dictionary learning problem (\ref{eq:2}) seek to: 
%
%\begin{itemize}
%
%\item Capture as much as possible information in the signal by minimizing the global reconstruction error. 
%\item Specialize the extracted information per class by minimizing the class specific reconstruction error similar to intra-class variations minimization. 
%\item Render dissimilar the extracted class specific information by promoting orthogonality of dictionaries and "zeroing" coefficients not specific to the sample label. In other words, we attempt to maximize inter-class variations.
%\item Promote coefficients sparsity to maintain generalization ability.
%\end{itemize}

\corGG{}{$\mu$}$, \lambda$, $\gamma_{1}$ and $\gamma_{2}$ are regularization parameters controlling \corGG{}{respectively the class specific fitting error}, the sparsity \corGG{}{level of each signal}, \corGG{the structure of sparse coefficients}{the sparsity structure of the codes} and pairwise orthogonality of learned dictionaries. \corGG{}{From this formulation we derive an optimization framework presented hereafter.}

%\textcolor{red}{We could have associated a regularization parameter to the term $J_{2}$, however to avoid multiplying the number of hyper-parameters we choose to fix it to $1$. Furthermore, conducted experiments show that it does not have significant impact on the performances. 
%}

\subsection{Optimization scheme}
\label{sec:opt}
\corGG{At the first sight, the objective function in (\ref{eq:2}) seems to be complex but it}{The optimization problem (\ref{eq:2}) may seem  complicated but it} can be solved based on an alternating optimization scheme which involves a sparse coding step and dictionary optimization step. Indeed, problem (\ref{eq:2}) is convex in $\mathbf{D}_{c}$ for the coefficients $\mathbf{a}_{nc}$ fixed and is so the \corGG{inverse}{reverse} way when the $\mathbf{D}_{c}$ are fixed.

\subsubsection{Sparse coding step}
\corGG{In this step, we fix $\{\mathbf {D}_{c}\}_{c=1}^{C}$ and we estimate the coefficients $\{\mathbf{a}_{n}\}_{n=1}^{N}$.}{Assume the dictionaries $\{\mathbf {D}_{c}\}_{c=1}^{C}$ are fixed; we estimate the sparse codes $\{\mathbf{a}_{n}\}_{n=1}^{N}$ using a Lasso-type algorithm \citep{lee2006efficient}. Minimizing $J(\D, \A)$ with relation to $\A$ amounts to minimize $J_{1}(\D, \A) + \mu J_{2}(\D, \A) + \lambda J_{3}(\A) +\gamma_{1} J_{4}(\A)$  over $\A$ as the other terms in $J$ are independent of $\A$. Moreover} for each signal $\mathbf {x}_{n}$ of class $y_{n}$, the related vector $\mathbf{a}_{n}$ is decoupled in the optimization problem.  Let $y_{n}=c'$; \corGG{this conducts us to solve the following problem: }{by putting apart all terms that do not involve $\mathbf{a}_n$, we are to solve the following optimization problem to estimate $\mathbf{a}_n$:} 
\begin{equation}
\displaystyle  \min_{\mathbf{a}_{n}}  \norme{\mathbf{x}_{n}-\mathbf{D} \mathbf{a}_{n}}_{2}^{2} +  \mu \norme{\mathbf{x}_{n}-\mathbf{D}_{c'} \mathbf{a}_{nc'}}_{2}^{2} + \gamma_{1} (\norme {\mathbf{a}_{n}}^{2}_{2} - \norme {\mathbf{a}_{nc'}}^{2}_{2}) + \lambda \norme {\mathbf{a}_{n}}_{1}
\label{eq:3}
 \end{equation}
where $ \displaystyle \norme{\mathbf{a}_{n}}_{2}^{2}= \sum_{c=1}^{C} \norme{\mathbf{a}_{nc}}_{2}^{2}$ and $\displaystyle \sum_{c=1}^{C} \un_{c \neq c'} \norme {\mathbf{a}_{nc}}_{2}^{2}= \norme{\mathbf{a}_{n}}_{2}^{2}- \norme{\mathbf{a}_{nc'}}_{2}^{2}$ \\

It can be seen that (\ref {eq:3}) consists of quadratic error terms and elastic-net type penalization \corGG{}{($\ell_1 - \ell_2$ norm penalty)}. Thus this problem is amenable to a Lasso problem which can be solved by a classical Lasso solver \citep{lee2006efficient}.

\subsubsection{Dictionary optimization step}

Here we illustrate the estimation of $\{\mathbf{D}_{p}\}_{p=1}^{C}$ while fixing $\{\mathbf{a}_{n}\}_{n=1}^{N}$. \corGG{It can be seen that (\ref{eq:2}) involves quadratic terms with respect to the dictionaries. The derivative of the objective function with respect to $\mathbf{D}_{p}$ is:}{Optimizing $J$ w.r.t the dictionaries $\mathbf{D}_{p}$ is equivalent to solve $\min_{\{\mathbf{D}_{p}\}_{p=1}^{C}} J_{1}(\D, \A) + \mu J_{2}(\D, \A) + \gamma_{2} J_{5}(\D)$ under the constraints $\norme{\mathbf{d}_{pk}}_{2}^{2} \leq 1, \forall p, k$. As the objective functions $J_1$, $J_2$ and $J_5$ are all quadratic with respect to the $\D_p$ and the constraints are simple, we adopt a gradient projection approach \cite{bertsekas1999nonlinear}. it  consists to update iteratively the dictionaries by $\D^t = \mbox{Prox} \big( \D^{t-1} - \eta_t \nabla J(\D^{t-1}, \A^t) \big)$ that is taking a gradient step followed by a projection onto the constraints via the proximal projection operator $\mbox{Prox}$ (see Algorithm \ref{alg:1}). This requires the computation of the gradient of the objective function with respect to $\mathbf{D}_{p}$ which is defined as follows:}  
\begin{equation}
 \nabla_{\mathbf{D}_{p}} J(\D, \A)= \nabla_{\mathbf{D}_{p}} J_{1}(\D, \A) + \mu  \nabla_{\mathbf{D}_{p}} J_{2}(\D, \A)+ \gamma_{2} \nabla_{\mathbf{D}_{p}} J_{5}(\D)  
\label{eq:der}
 \end{equation}
The involved terms are obtained below using the matrix derivation formula \citep{petersen2008matrix}. \corGG{}{Notice that $J_{1}(\D, \A)=\displaystyle \sum_{n=1}^{N} \norme{\mathbf{x}_{n}-\mathbf{D} \mathbf{a}_{n}}_{2}^{2}$ can also take the form $J_{1}(\D, \A) = \sum_{n=1}^{N} \norme{ \tilde{\mathbf{x}}_{n}-\mathbf{D}_{p} \mathbf{a}_{np}}_{2}^{2}$ where $\displaystyle \tilde{\mathbf{x}}_{n}=\mathbf{x}_{n}-\sum_{\substack {c=1 \\ c \neq p}}^{C} \mathbf{D}_{c} \mathbf{a}_{nc}$. Hence the derivative is}
\begin{equation}
 %\left\{
  %  \begin{array}{ll}
%J_{1}(\D, \A)=\displaystyle \sum_{n=1}^{N} \norme{\mathbf{x}_{n}-\mathbf{D} \mathbf{a}_{n}}_{2}^{2}= \sum_{n=1}^{N} \norme{ \tilde{\mathbf{x}}_{n}-\mathbf{D}_{p} \mathbf{a}_{np}}_{2}^{2} \\ % \hspace{5mm} \tilde{\mathbf{x}}_{n}=\mathbf{x}_{n}-\sum_{\substack {c=1 \\ c \neq p}}^{C} \mathbf{D}_{c} \mathbf{a}_{nc} \\
% & \\
\nabla_{\mathbf{D}_{p}} J_{1}(\D, \A)= \sum_{n=1}^{N} -2 \mathbf{\tilde{x}}_{n} \mathbf{a}_{np}^{T}+2 \mathbf{D}_{p} \mathbf{a}_{np} \mathbf{a}_{np}^{T}
%\end{array}
%\right.
\label{eqq1}
\end{equation}
\noindent \corGG{For the second term of the derivative  $\nabla_{\mathbf{D}_{p}} J$ we can write}{Similarly we can express the term $J_2$ as $J_{2}(\D, \A)= \sum_{n=1}^{N}  \un_{y_{n} = p}  \norme{\mathbf{x}_{n}-\mathbf{D}_{p} \mathbf{a}_{np}}_{2}^{2} + \sum_{n=1}^{N} \sum_{c \neq p}   \un_{y_{n}=c} \norme{\mathbf{x}_{n} - \mathbf{D}_{c} \mathbf{a}_{nc}}_{2}^{2}$. Hence the second term of the gradient writes}

\begin{equation}
% \left\{
 %   \begin{array}{ll}
%J_{2}=\displaystyle \sum_{n=1}^{N}  \un_{y_{n} = p}  \norme{\mathbf{x}_{n}-\mathbf{D}_{p} \mathbf{a}_{np}}_{2}^{2} + \sum_{n=1}^{N} \sum_{c \neq p}   \un_{y_{n}=c} \norme{\mathbf{x}_{n} - \mathbf{D}_{c} \mathbf{a}_{nc}}_{2}^{2}                \\
%& \\
\nabla_{\mathbf{D}_{p}} J_{2}= \displaystyle \sum_{n=1}^{N}  \un_{y_{n} = p} -2 \mathbf{x}_{n} \mathbf{a}_{np}^{T}+2 \mathbf{D}_{p} \mathbf{a}_{np} \mathbf{a}_{np}^{T} 
%\end{array}
%\right.
\end{equation}

\noindent \corGG{Finally the expression of the last term is given by}{Finally expressing $J_{5}(\D)=\displaystyle \sum_{c \neq p} 2 \norme {\mathbf{D}_{p}^{T} \mathbf{D}_{c}}_{F}^{2} + \sum_{c \neq p} \sum_{\substack {c' \neq c \\ c' \neq p}} \norme {\mathbf{D}_{c}^{T} \mathbf{D}_{c'}}_{F}^{2}$ we get the last term of the gradient as }

\begin{equation}
% \left\{
 %   \begin{array}{ll}
%J_{5}=\displaystyle \sum_{c \neq p} 2 \norme {\mathbf{D}_{p}^{T} \mathbf{D}_{c}}_{F}^{2} + \sum_{c \neq p} \sum_{\substack {c' \neq c \\ c' \neq p}} \norme {\mathbf{D}_{c}^{T} \mathbf{D}_{c'}}_{F}^{2} \\
%& \\
\nabla_{\mathbf{D}_{p}} J_{5}(\D)= \displaystyle \sum_{c \neq p} 4 (\mathbf{D}_{c} \mathbf{D}_{c}^{T}) \mathbf{D}_{p}
%\end{array}
%\right.
\label{eqq2}
\end{equation}

\vspace{0.5cm}

%\begin{equation}
%\nabla_{\mathbf{D}_{p}} J = \sum_{n=1}^{N} -2 \tilde{\mathbf{x}}_{n} \mathbf{a}_{np}+ 2 \mathbf{D}_{p} \mathbf{a}_{np} \mathbf{a}_{np}^{T} + \gamma_{1} \sum_{n=1}^{N} \un_{y_{n} \neq p} (-2 \mathbf{a}_{n} \mathbf{a}_{np} + 2  \mathbf{D}_{p} \mathbf{a}_{np} \mathbf{a}_{np}^{T})+ \gamma_{3} \sum_{\substack {c=1 \\ c \neq p}}^{C} 4 (\mathbf{D}_{c} \mathbf{D}_{c}^{T}) \mathbf {D}_{p}
%\end{equation}

Algorithm \ref {alg:1} summarizes the different steps \corGG{of our optimization approach which is based on an alternating scheme:}{of our alternating optimization scheme:} the first step consists of a signal sparse coding based on the Lasso algorithm. The second step is dictionary optimization based on proximal gradient descent approach. The proximal procedure \corGG{is useful in order}{allows} to handle the atom normalization constraint $\norme{\mathbf{d}_{ck}}_2^2 \leq 1$ in the problem (\ref{eq:2}).

\begin{algorithm} [!h]
\begin{algorithmic} [1]
\STATE\textbf{ Initialization:}  $\mathbf {D}^{0}$, $t \leftarrow 1$, \mbox{initialize} $\eta_{0}$ and $\alpha \in (0, 1)$
\WHILE {$t \leq T$}
\STATE \mbox{Solve for} $\mathbf{A}^{t} \leftarrow \displaystyle \argmin_{\mathbf{A}} J(\mathbf{D}^{t-1},\mathbf{A})$ \mbox{using Lasso algorithm} \corGG{}{applied to (\ref{eq:3})}\\
\STATE \mbox{Compute the gradient} $ \mathbf{G}^{t}=\nabla_{\mathbf{D}} J(\mathbf{D}^{t-1}, \mathbf{A}^{t})$ \mbox{based on eq. (\ref{eq:der}) to (\ref{eqq2})} \\
\STATE $\eta \leftarrow \eta_{0}$ 
\REPEAT
\STATE $ \mathbf{D}^{\frac{t}{2}} \leftarrow \mathbf{D}^{t-1} - \eta \, \mathbf{G}^{t}$ \\
\STATE $\mathbf{D}^{t} \leftarrow \mbox{Prox} \big(\mathbf{D}^{\frac{t}{2}}\big)$ \\

\begin{equation*}
 \mbox{with} \hspace {0.8cm} \mbox{Prox} \big(\mathbf{D}^{\frac{t}{2}}\big) :   \{\mathbf{d}_{k}\}_{k=1}^{K}= \left\{
    \begin{array}{l}
   \mathbf{d}_{k} \hspace{8mm} \mbox{if}   \hspace{4mm} \norme{\mathbf{d}_{k}}_{2} \leq 1 \\
        \\
       \frac{\mathbf{d}_{k}}{\norme{\mathbf{d}_{k}}_{2}} \hspace{4mm} \mbox{otherwise} 
    \end{array}\right.
    \end{equation*}
    \STATE $\eta \leftarrow \eta \times \alpha$
\UNTIL {$J(\mathbf{D}^{t},\mathbf{A}^{t})<J(\mathbf{D}^{t-1},\mathbf{A}^{t-1})$}
\STATE $t \leftarrow t+1$
\ENDWHILE
\end{algorithmic}
\caption{The optimization algorithm}
\label{alg:1}
\end{algorithm}

\subsection{Classification}

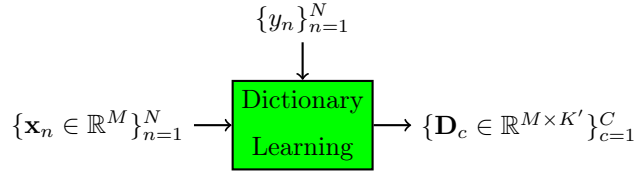
\begin{figure}[!h]
\centering
\begin{tikzpicture}  [thick,scale=1, every node/.style={scale=1}]
\tikzstyle{box} = [rectangle,draw,thick,align=center,minimum height=10mm];
\tikzstyle{arrow} = [->,thick];

\node[] (d) {$\{\mathbf{x}_{n} \in \mathbb{R}^{M}\}_{n=1}^{N}$};

\node[box,right=5mm of d.east,anchor=west,fill=green] (tf) {Dictionary\\ Learning};

\node[right=5mm of tf.east,anchor=west] (dicolearn) {$\{\mathbf{D}_{c} \in \mathbb{R}^{M \times K'}\}_{c=1}^{C}$};

\node[above=5mm of tf.north,anchor=south] (target) {$\{y_{n}\}_{n=1}^{N}$};
\draw[arrow] (target)--(tf);

\draw[arrow] (d)--(tf);
\draw[arrow] (tf)--(dicolearn);

\end{tikzpicture}
\caption{Processing flow of dictionary learning on the training set.} \label{fig:feature_selection_contribution1}
\end{figure}

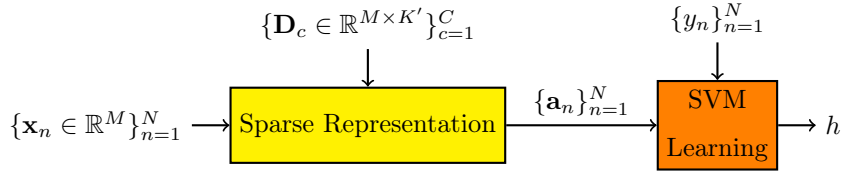
\begin{figure}[!h]
\centering
\begin{tikzpicture}  [thick,scale=1, every node/.style={scale=1}]
\tikzstyle{box} = [rectangle,draw,thick,align=center,minimum height=10mm];
\tikzstyle{arrow} = [->,thick];

\node[] (d) {$\{\mathbf{x}_{n} \in \mathbb{R}^{M}\}_{n=1}^{N}$};

\node[box,right=5mm of d.east,anchor=west,fill=yellow] (dico) {Sparse Representation};

\node[above=5mm of dico.north,anchor=south] (dicolearn) {$\{\mathbf{D}_{c} \in \mathbb{R}^{M \times K'}\}_{c=1}^{C}$};

\node[box,right=20mm of dico.east,anchor=west,fill=orange] (svm) {SVM\\ Learning};

\node[above=5mm of svm.north,anchor=south] (target) {$\{y_{n}\}_{n=1}^{N}$};

\node[right=5mm of svm.east,anchor=west] (dddd) {$h$};

\draw[arrow] (d)--(dico);
\draw[arrow] (dicolearn)--(dico);
\draw[arrow] (dico)--(svm) node[above,pos=0.5] {$\{\mathbf{a}_{n}\}_{n=1}^{N}$};
\draw[arrow] (svm)--(dddd);
\draw[arrow] (target)--(svm);

\end{tikzpicture}
\caption{Processing flow of SVM training over the learned dictionary and training set.} \label{fig:feature_selection_contribution2}
\end{figure}

\begin{figure}[!h]
\centering
\begin{tikzpicture}  [thick,scale=1, every node/.style={scale=1}]
\tikzstyle{box} = [rectangle,draw,thick,align=center,minimum height=10mm];
\tikzstyle{arrow} = [->,thick];

\node[] (d) {$\{\mathbf{x}_{n'} \in \mathbb{R}^{M}\}_{n'=1}^{N_{test}}$};

\node[box,right=5mm of d.east,anchor=west,fill=yellow] (dico) {Sparse Representation};

\node[above=5mm of dico.north,anchor=south] (dicolearn) {$\{\mathbf{D}_{c} \in \mathbb{R}^{M \times K'}\}_{c=1}^{C}$};

\node[box,right=20mm of dico.east,anchor=west,fill=cyan] (svm) {SVM\\ Classification};

\node[above=5mm of svm.north,anchor=south] (svmlearn) {$h$};

\node[right=5mm of svm.east,anchor=west] (dddd) {$\{\tilde{y}_{n'}\}_{n'=1}^{N_{test}}$};

\draw[arrow] (d)--(dico);
\draw[arrow] (dicolearn)--(dico);
\draw[arrow] (dico)--(svm) node[above,pos=0.5] {$\{\mathbf{a}_{n'}\}_{n'=1}^{N_{test}}$};
\draw[arrow] (svm)--(dddd);
\draw[arrow] (svmlearn)--(svm);

\end{tikzpicture}
\caption{Processing flow of classification over testing set.} \label{fig:feature_selection_contribution3}
\end{figure}
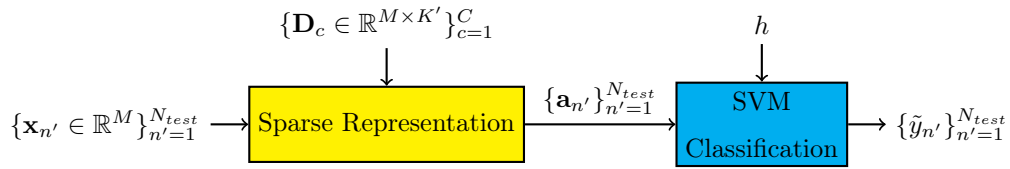

\corGG{Once the dictionaries are learned, they are used to encode both training and testing samples based on Lasso. The resulting coefficients are used to feed an SVM classifier. Figures \ref{fig:feature_selection_contribution1} to \ref{fig:feature_selection_contribution3} show the processing flow of dictionary learning based on the training data, coding both training and testing data over the learned dictionary respectively.}{Our overall signal classification scheme consists of the following steps:
\begin{itemize}
	\item[(i)] the class specific dictionaries $\{\mathbf{D}_{c}\}_{c=1}^{C}$ are estimated in a supervised way using Algorithm \ref{alg:1} as shown in figure \ref{fig:feature_selection_contribution1};
	\item[(ii)] the dictionaries are then used to encode the training signals (based on Lasso), leading to the sparse codes $\{ \ba_n \}_{n=1}^N$ which serve as features to learn an SVM function $h$. This is summarized by the processing flow in figure \ref{fig:feature_selection_contribution2}; and
	\item[(iii)] any testing signal is classified by computing its sparse representation which is fed to the classifier $h$ to predict the corresponding label (see figure \ref{fig:feature_selection_contribution3}).
\end{itemize}
}

To solve our $C$-class audio classification problem we employ one-against-all strategy \cite{scholkopf2002learning}.
%It consists in constructing $C$ binary SVM, each one separates a class from all the rest.
%The $c^{th}$ SVM solves the decision problem $h^{(c)}(\mathbf{a})=h_{0}^{(c)}(\mathbf{a})+b ^ {(c)}$ with data from class $c$ taken as positive samples and the remaining training samples as negatives. 
Note that in our case we have used a simple linear kernel as the non-linear aspect of the problem is taken into account in the dictionary learning. This is customary in supervised dictionary classification \citep{mairal2009,mairal2012task}.

\section{Experiments}
\label{exp}
We conduct our experiments on two different audio signal classification problems, Computational Auditory Scene Recognition (CASR) and music chord recognition. For each problem, our dictionary learning based on a initial time-frequency representation is compared to conventional \corGG{hand-crafted}{predefined}  features.

\subsection{Computational auditory scene recognition (CASR)}
In this section we briefly review different approaches to tackle CASR problem as well as the evaluation of our proposed dictionary learning technique compared with \corGG{hand-crafted}{predefined} features based approaches on two datasets: East Anglia (EA) and LITIS Rouen.

Several categories of audio features have been employed in CASR systems \citep{barchiesi2015acoustic}. A considerable amount of works have applied MFCCs for CASR. Aucouturier \textit{et al.} \citep {aucouturier2007bag} used Gaussian Mixture Model (GMM) to estimate the distribution of MFCC coefficients. Ma \textit{et al.} \citep {ma2006acoustic} combined MFCCs with Hidden Markov Models (HMM). Cauchi \citep{cauchi2011} exploited Non-Negative Matrix Factorization (NMF) with MFCC features. Hu \textit{et al.} \citep{hu2012combining} employed MFCC features in a two-stage framework based on GMM and SVM. Lee \textit{et al.} \citep{lee2013acoustic} used sparse restricted Boltzmann machine to capture relevant MFCC coefficients. Geiger \textit{et al.} \citep{geiger2013large} extracted a large set of features including MFCCs using a short sliding window approach. SVM is used to classify these short segments, and a majority voting scheme is employed for the whole sequence decision. Roma \textit{et al.} \citep {roma2013recurrence} applied Recurrence Quantification Analysis (RQA) on the MFCCs for supplying some additional information on temporal dynamics of the signal.

Another trend is to extract discriminative features from time-frequency representations. Cotton and Ellis \citep{cotton2011spectral} applied NMF to extract time-frequency patches. Benetos \textit{et al.} \citep{benetos2012characterisation} used temporally-constrained Shift-Invariant Probabilistic Latent Component Analysis (SIPLCA) instead of NMF in order to extract time-frequency patches from spectrogram. Yu and Slotine \citep{yu2008audio} proposed a method based on treating time-frequency representations of audio signals as image texture. In the same context, Dennis \textit{et al.} \citep{dennis2013image} introduced novel sound event image representation called Subband Power Distribution (SPD). The SPD captures the distribution of the sound's log-spectral power over time in each subband. Rakotomamonjy and Gasso \citep{Rakotomamonjy:2015} proposed to use Histogram of Oriented Gradient to extract information from time-frequency representations.

\subsubsection {Datasets}
We rely our experiments on two representative datasets described hereafter.
\begin{itemize} 
\item East Anglia (EA): this dataset \footnote{\url{http://lemur.cmp.uea.ac.uk/Research/noise_db/}}
 provides environmental sounds \citep {ma2003context} coming from 10 different locations: \textit{bar, beach, bus, car, football match, launderette, lecture, office, rail station, street}. In each location a recording of 4-minutes at a frequency of 22.1 kHz has been collected. The 4-minutes recordings are splitted into 8 recordings of 30-seconds so that in total we have 10 locations (classes) and each class has 8 examples of 30-seconds. 
\end{itemize} 

\begin{itemize} 
\item Litis Rouen: this dataset \footnote{\url{https://sites.google.com/site/alainrakotomamonjy/home/audio-scene}} provides environmental sounds  \citep{Rakotomamonjy:2015} recorded in 19 locations. Each location has different number of 30-seconds examples downsampled at 22.5 kHz. Table \ref{tab:liti} summarizes the content of the dataset.
\end{itemize}

\begin{table*} [!h] \centering
 \caption{Summary of Litis Rouen audio scene dataset.}
\scalebox{0.7}{
\begin{tabularx}{11.5cm}{ Xc}
\hline
\hline
\textbf{Classes} &  \hspace{2mm} \textbf{\# examples} \\
\hline
\hline
plane  & 23  \\
busy street  & 143 \\
bus & 192\\
cafe       & 120    \\
car       &  243  \\
train station hall   & 269  \\
kid game hall    & 145   \\
market       &   276 \\
metro-paris       & 139  \\
metro-rouen       & 249   \\
billiard pool hall       & 155  \\
quite-street       &    90 \\
student hall       &  88  \\
restaurant      &   133 \\
pedestrian street       & 122   \\
shop      & 203    \\
train       &  164  \\
high-speed train  &  147 \\
tube station       &   125  \\
\hline
\hline
\end{tabularx}}
\label{tab:liti}
\end{table*}

\subsubsection {Competing features and protocols}
In the following we introduce the different features used in our experiments as well as the data partition and protocols.

\subsubsection*{Features}
Based on an initial time-frequency representation (spectrogram) computed on sliding windows of size $4096$ samples and hops of $32$ samples, we apply our dictionary learning method. In order to evaluate the efficiency of our proposed method, we compare its performance to the following conventional features:

\begin {itemize}

%\item Spectrogram pooling: represents the temporal pooling of the spectrogram computed on sliding windows of size $4096$ samples and hops of $32$ samples.
\item Bag of MFCC: consists in calculating the MFCC features on windows of size $25$ ms with hops of $10$ ms. For each window, $13$ cepstra over $40$ bands are computed (lower and upper band are set to $1$ and $10$ kHz). The final feature vector is obtained by concatenating the average and standard deviation of the batch of $40$ windows with overlap of $20$ windows.
\item Bag of MFCC-D-DD: in addition to the average and standard deviation, the first-order and second-order differences of the MFCC over the windows are concatenated to the feature vector. 
\item Texture-based time-frequency representation: it consists on extracting features from time-frequency texture \citep{yu2008audio}.
\item Recurrent Quantification Analysis (RQA): aims to extract from MFCCs some additional information on temporal dynamics. For all MFCCs obtained over $40$ windows with overlap of $20$, $11$ RQA features have been computed \citep {roma2013recurrence}. Afterwards, MFCC features and RQA features are all averaged over time and MFCC averages, standard deviations as well as the RQA averages are concatenated to form the final feature vector.
\item HOG of time-frequency representation: applies HOG to time-frequency representations transformed to images. The time-frequency representations are calculated based on Constant-Q Transform (CQT). HOG is able to provide information about the occurrence of gradient orientations in the resulting images \citep{Rakotomamonjy:2015}.
\end{itemize}
More details about these features can be found in  \citep{Rakotomamonjy:2015}. Note that for classification, linear Support Vector Machine (SVM) is applied.

\subsubsection*{Protocols and parameters tuning}
For sake of comparison we have performed the same experiments using the same repartitions and protocols in \citep{Rakotomamonjy:2015}. We have averaged the performances from $20$ different  splits of the initial data into training and test. The training set represents $80$ \% of data while the rest represents the test set. 

\corGG{Our proposed dictionary learning technique requires the following parameters:}{Our proposed dictionary learning technique requires the tuning of some hyper-parameters: $K'$ the size of each dictionary $\mathbf{D}_{c}$, $\lambda$, $\gamma_{1}$, $\gamma_{2}$ controlling respectively, the sparsity, the structure of sparse coefficients and the pairwise orthogonality of learned dictionaries and and $\mu$ the weight affected to the class specific reconstruction error $J_2$. To avoid a tedious hyper-parameters' selection step and guided by empirical findings, we fix $\mu=1$. Hence the remaining parameters are determined as follows:}
%We could have associated a regularization parameter to the term $J_{2}$, however to avoid multiplying the number of hyper-parameters we choose to fix it to $1$. Furthermore, conducted experiments show that it does not have significant impact on the performances. 

\begin {itemize}
\item $\lambda$, $\gamma_{1}$ and $\gamma_{2}$ \removeGG{controlling respectively, the sparsity, the structure of sparse coefficients and pairwise orthogonality of learned dictionaries. The parameters} are selected among $\{0.1, 0.2, 0.3\}$.
\item  \corGG{$K^{'}$ the size of each dictionary $\mathbf{D}_{c}$. Its value is explored among $\{10, 20, 30\}$.}{the size $K^{'}$ of each dictionary is explored among $\{10, 20, 30\}$.}
\end{itemize}

Beyond that we use a linear SVM classifier which regularization parameter $C_{svm}$ is selected among $10$ values  logarithmically scaled between $0.001$ and $100$. All these parameters are tuned according to a cross-validation scheme. Model selection is performed by resampling $5$ times the training set into learning and validation sets of equal size. The best parameters are considered as those maximizing the averaged performances on the validation sets. Note that K-SVD  \citep{aharon2006img} has been used to initialize the class based dictionaries and the parameters $T=200$,  $\alpha=0.5$ and $\eta=10^{-3}$ were applied for the optimization scheme (see Section \ref{sec:opt}).

\subsubsection{Results and analysis}

Table \ref{tab_ comp_casr} represents the performance (classification accuracy) comparison between different conventional features as reported in \citep{Rakotomamonjy:2015} and our class based dictionary method on Rouen and EA datasets. Texture denotes the work of \citep{yu2008audio} while MFCC-D-DD denotes the MFCC with derivatives features. MFCC, MFCC-RQA, MFCC-900 and MFCC-RQA-900 respectively denote, MFCC features, the MFCC with RQA with cut-off frequency of 10 kHz, the MFCC and the MFCC combined RQA with upper frequency set at 900 Hz respectively. HOG-full and HOG-marginalized represent the concatenation of histogram obtained from different cells resulting in a very-high dimensionality feature vector and the concatenation of the averaged histograms over time and frequency respectively.

\begin{table*} [!h] \centering
 \caption{Comparison of performances related to different feature representations on Rouen, EA  audio scene classification datasets. Bold values stand for best values on each dataset.}
\scalebox{0.8}{
\begin{tabularx}{15cm}{ X X l}
\hline
\hline
\textbf{Features} &  \hspace{2mm} \textbf{Rouen} &    \hspace {5mm} \textbf {EA} \\
\hline
\hline
Texture       &   \hspace{8mm} -  &   0.57 $\pm$ 0.13 \\
\hline
\hline
MFCC-D-DD  &  $0.66$ $\pm$ $0.02$ &   0.98 $\pm$ 0.04  \\
\hline
\hline
MFCC  &  0.67 $\pm$ 0.01  &   \textbf{1.00} $\pm$ \textbf{0.01}  \\
MFCC-900  &  0.60 $\pm$ 0.02 &   0.91 $\pm$ 0.07  \\
\hline
\hline
MFCC+RQA &  0.78 $\pm$ 0.01  &   0.95 $\pm$ 0.08 \\
MFCC+RQA-900 &  0.72 $\pm$ 0.02   &   0.93 $\pm$ 0.06  \\
\hline
\hline
HOG-full &   0.84 $\pm$ 0.01   &   0.99 $\pm$ 0.02   \\
HOG-marginalized &   \textbf{0.86} $\pm$ \textbf{0.01}   &   0.97 $\pm$ 0.06  \\
\hline
\hline
Dictionary learning &   0.71 $\pm$ 0.01   &   0.97 $\pm$ 0.04  \\
\hline
\hline
\end{tabularx}}
\label {tab_ comp_casr}
\end{table*}

%\begin{figure}[!h]
%\centering
%\includegraphics [width= 12 cm] {dictionary_rouen_new.pdf}
%\caption{Example of learned dictionaries per class on Rouen dataset. Rows correspond to learned dictionary atoms.}
%\label{fig:rouendictionaries}
%\end {figure}

\begin{figure}[!h]
\centering
\includegraphics [width= 7 cm] {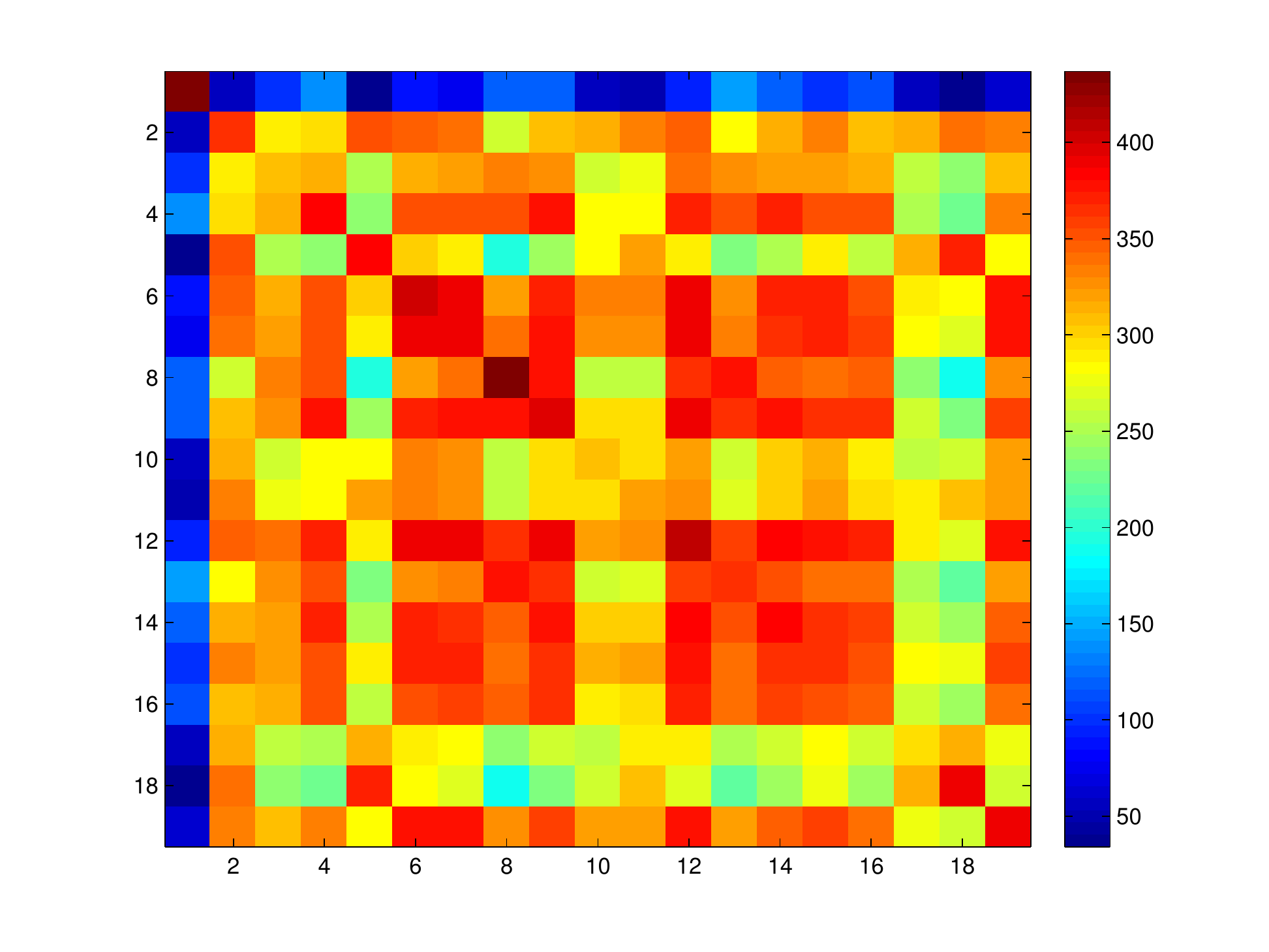}
\caption{Similarity between different learned dictionaries on Rouen dataset. X-axis and Y-axis stand for the class numbers organized in the same order in Table \ref{tab:liti}.}
\label{fig:rouensimilarity}
\end {figure}

It can be seen in Table \ref{tab_ comp_casr} that HOG-marginalized outperforms all competing features in Rouen dataset. Note also that MFCC+RQA features are performing better than other MFCC based features, however the cut-off-frequency of $900$ Hz leads to a large loss in performance. We can also notice that our proposed dictionary learning is giving very promising results and is outperforming texture and conventional speech recognition feature, MFCC and MFCC-D-DD features which have been widely used in the literature and have showed their ability to tackle the problems of audio scene recognition. Finally, in the East Anglia dataset, all features including our proposed dictionary learning perform well except texture, however we should note a slight advantage of MFCC.

 Figure \ref{fig:rouensimilarity} shows the pairwise similarity of the learned dictionaries per class on Rouen dataset. The idea behind estimating the similarity between different learned dictionaries is to verify the initial goal to learn dissimilar dictionaries able to extract diverse information from classes for discrimination purpose. It can be seen that there is some similarity between some learned dictionaries which could influence the classification accuracy since these dictionaries tend to provide similar information for different classes. This may be related to the increasing number of classes that makes enforcing the pairwise dictionaries dissimilarity hardly feasible.

\subsection {Music chord recognition}
The simplest definition of a chord is few musical notes played at the same time. In western music, each chord can be characterized by the:

\begin{itemize}
 
 \item  \textit {root or fundamental:}  the fundamental note on which the chord is built
 \item   \textit{number of notes}
 \item \textit {type}: gives the interval scheme between notes
  
\end{itemize}

 A music signal can be deemed composed of sequences of these different chords. Commonly, the duration of the chords in the sequence varies over time  rendering  their recognition difficult. Given a raw audio signal, chord recognition system attempts to automatically determine the sequence of chords describing the harmonic information. To recognize chords most approaches rely on features crafted based on time-frequency representation of the raw signals, the most common and dominant features being chroma \citep{oudre2009template}. Pitch Class Profiles (PCP) or chroma vectors was introduced by Fujishima \citep {fujichroma}. It is a 12-dimensional vectors representing the energy within an equal-tempered chromatic scale $ \{C , C^ {\#} , D ,\cdots, B \}$. The chroma has several variations, among them we can cite Harmonic Pitch Class Profiles (HPCPs) which is an extension of the Pitch Class Profiles (PCPs) by estimating the harmonics \citep {papadopoulos2008simultaneous} and Enhanced Pitch Class Profile (EPCP) which is calculated using the harmonic product spectrum \citep{lee2006automatic}. Chroma vectors were combined with different machine learning techniques \citep{sheh2003chord,weller2009structured}.

\subsubsection {Dataset}

We will focus on third, triad and seventh chords which are respectively composed of 2, 3 and 4 notes.
When a note B has twice the frequency of a note A, the interval $[A\;B]$ forms an octave.
In tempered occidental music, the smallest subdivision of an octave is a semitone which corresponds to one twelfth of an octave, that is a multiplication by $\sqrt[12]{2}$ in term of frequency.
To be tertian, i.e a standard harmony, each interval between notes in a chord must be composed of 3 or 4 semitones.These intervals are respectively called \textit{minor} and \textit{Major}.
Thus, for a given root, there is 2 possible thirds, 4 possible triads, and 8 possible sevenths.
Table \ref{tab:typeofchords} sum-up all the possible tertian third, triad and seventh chords. The pursued goal in this work is to guess the type and not the fundamental of a chord  leading to 14 possible labels ($=2+4+8$).  For this purpose, we have created a dataset which contains 2156 music chord samples of duration $2$-seconds at frequency $44100$~Hz with the 14 different classes.
Each class contains 154 samples from 11 different instruments at different fundamentals.

\begin{table}[!ht] \centering
\caption{Different kind of tertian chords, intervals are in semitones}
\label{tab:typeofchords}
\scalebox{0.7}{
\begin{tabular}{ccccc}
\hline
\hline
\# of notes & Common name or type & 1st interval &  2nd int. & 3rd int.\\
\hline
\hline
2 & Minor third & 3 & - & -\\
2 & Major third & 4 & - & -\\
\hline
3 & Diminished triad& 3 & 3 & -\\
3 & Minor triad& 3 & 4 & -\\
3 & Major triad& 4 & 3 & -\\
3 & Augmented triad& 4 & 4 & -\\
\hline
4 & Diminished seventh& 3 & 3 & 3\\
4 & Half-diminished seventh& 3 & 3 & 4\\
4 & Minor seventh& 3 & 4 & 3\\
4 & Minor major seventh& 3 & 4 & 4\\
4 & Dominant seventh& 4 & 3 & 3\\
4 & Major seventh& 4 & 3 & 4\\
4 & Augmented major seventh& 4 & 4 & 3\\
4 & Augmented augmented seventh & 4 & 4 & 4\\
\hline
\hline
\end{tabular}}
\end{table}

\subsubsection {Competing features and protocols}

In the following we introduce the different features used in our experiments as well as the data partition and protocols.

\subsubsection*{Features}
Similar to the previous application we compute an initial time-frequency representation (spectrogram) on sliding windows of size 4096 samples and hops of 32 samples. Then we apply our dictionary learning method. The resulting sparse representations are used as inputs of an SVM. The following conventional features serve as competitors to our approach.

\begin {itemize}
\item Spectrogram pooling: represents the temporal pooling of the spectrogram.
\item Interpolated power spectral density: music notes follow an exponential scale, however Power Spectral Density (PSD) is based on Fourier transform which follows a linear scale. To address this problem PSD (which lies on a linear scale) is sampled at specific frequencies corresponding to 96 notes leading to an exponential representation more suitable for chord recognition \citep{rida2014supervised}.
\item Chroma: it represents a $12$-dimensional vector, every component represents the spectral energy of a semi-tone within the chromatic scale. Chroma vector entries are calculated by summing the spectral density corresponding to frequencies belonging to the same chroma \citep{oudre2009template}.
\end{itemize}

\subsubsection*{Protocols and parameters tuning}

We have averaged the performances from different 10 splits of the initial data into training and test. The training set represents 2/3 of data. Model selection is performed by resampling $2$ times the training set into learning and validation set of equal size. The best parameters are considered as those maximizing the averaged performances on the validation sets. Note that the parameters are chosen from the same intervals used above in the computational auditory scene recognition problem.

\subsubsection{Results and analysis}

Table \ref{tab:tab_ comp_chord} reports the performance (classification accuracy) comparison of evaluated features on music chord dataset. It can be seen that our dictionary learning method outperforms all other approaches. 

\begin{table*} [h] \centering
 \caption{Comparison of performances related to different feature representations on music chord dataset based on linear SVM. Bold value stands for best performance.}
\scalebox{0.9}{
\begin{tabularx}{10cm}{ Xc}
\hline
\hline
\textbf{Features} &  \hspace{2mm} \textbf{Music chord} \\
\hline
\hline
Chroma  & 0.19 $\pm$ 0.01 \\
Interpolated PSD  & 0.15 $\pm$ 0.02 \\
Spectrogram pooling &  0.14 $\pm$ 0.01\\
Dictionary learning       &    \textbf{0.66 $\pm$ 0.01}  \\
\hline
\hline
\end{tabularx}}
\label {tab:tab_ comp_chord}
\end{table*}

%Table \ref{tab:tab_ comp_chord_nonl} represents the performance (classification accuracy) comparison of evaluated features on music chord dataset based on the polynomial kernel. It can be seen the interpolated PSD outperforms chroma and spectrogram. It can be also noticed that the polynomial kernel overcome the linear one in this particular task of chord recognition based on the conventional hand-crafted features.

%\begin{table*} [h] \centering
 %\caption{Comparison of performances related to different feature representations on music chord dataset based on polynomial kernel. Bold value stands for best performance.}
%\scalebox{1}{
%\begin{tabularx}{10cm}{ Xc}
%\hline
%\hline
%\textbf{Features} &  \hspace{2mm} \textbf{Music chord} \\
%\hline
%\hline
%Chroma  & 0.70 $\pm$ 0.01 \\
%Interpolated PSD  & \textbf{0.74 $\pm$ 0.01} \\
%Spectrogram pooling &  0.72 $\pm$ 0.01\\
%\hline
%\hline
%\end{tabularx}}
%\label {tab:tab_ comp_chord_nonl}
%\end{table*}

%\begin{figure}[!h]
%\centering
%\includegraphics [width= 9 cm] {dic_chord.pdf}
%\caption{Example of learned dictionaries per each class on music chord dataset.}
%\label{fig:rouendictionariess}
%\end {figure}

\begin{figure}[!h]
\centering
\includegraphics [width= 6 cm] {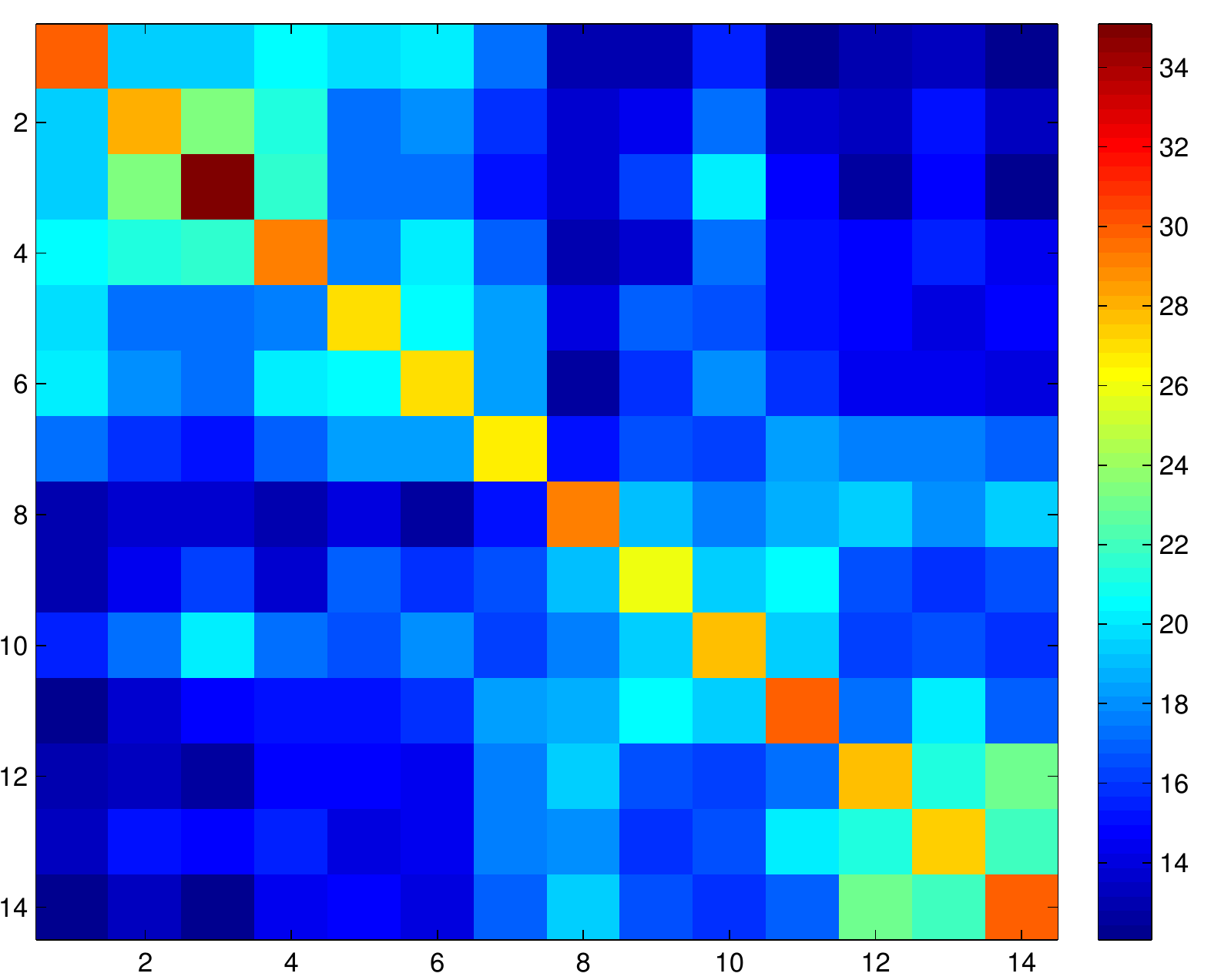}
\caption{Similarity between different learned dictionaries on music chord dataset. X-axis and Y-axis stand for the class numbers.}
\label{fig:rouensimilarityy}
\end {figure}

Figure \ref{fig:rouensimilarityy} shows the pairwise similarity between the learned dictionaries. Contrary to CASR Rouen dataset, it can be seen that the highest similarity between learned dictionaries is on the diagonal. This means that the  resulting dictionaries are different between them leading to extract diverse information per class. While chroma, interpolated PSD and spectrogram failed totally to reach good performances based on a linear SVM, our dictionary learning method could achieve very promising results. \corGG{Linear classification is a computationally efficient way to categorize test samples. It consists in finding a linear separator between two classes. Linear classification has been the focus of much research in machine learning for decades and the resulting algorithms are well understood. However, many datasets cannot be separated linearly and require complex nonlinear classifiers which is the case of our music chord dataset. A popular solution to enjoy the benefits of linear classifiers is to embed the data into a high dimensional feature space, where a linear classifier eventually exists. The feature space mapping is chosen to be nonlinear in order to convert nonlinear relations to linear relations. This nonlinear classification framework is at the heart of the popular kernel-based methods \citep{shawe2004kernel}. Despite the popularity of kernel-based classification, its computational complexity at test time strongly depends on the number of training samples \citep{burges1998tutorial}, which limits its applicability in large scale datasets. An eventual alternative to kernel methods, is sparse coding which consists in finding a compact representation of the data in an overcomplete learned dictionary which can be seen as a nonlinear feature representation mapping. This is confirmed by our experiments which clearly shows that our proposed dictionary learning method outperforms the other hand-crafted features. }{As a conclusion, the sparse coding of the signals over the learned dictionaries can be seen as a nonlinear feature mapping which is able to disentangle the factors of variation within the audio samples of different labels.}

\section{Conclusion}
\label{conc}
We have proposed a novel supervised dictionary learning method for audio signal recognition. The proposed method seeks to minimize the intra-class homogeneity, maximize the class separability and promote the sparsity to control the complexity of the signal decomposition over the dictionary. This is done by learning a dictionary per class, minimizing the class based reconstruction error and promoting the pairwise orthogonality of the dictionaries. The learned dictionaries are supposed to provide \corGG{diverse}{different} information per class. The resulting problem is non-convex and solved using a proximal gradient descent method.

Our proposed method was extensively tested on two different audio recognition applications: computational auditory scene recognition and music chord recognition. The obtained results were  compared to different conventional \corGG{hand-crafted}{predefined} features.  While there is no universal \corGG{hand-crafted}{pre-specified} feature representation able to successfully tackle different audio recognition problems, our proposed dictionary learning method combined with a simple linear classifier showed very promising results while dealing with \corGG{the two diverse recognition problems}{two different audio recognition tasks.}

%Despite the simplicity and good performances of our approach, we could notice that the task to make the learned dictionaries as different as possible is hardly feasible when dealing with large number of classes. An example is human identity recognition based on gait where each individual is seen as a class. 

%A possible alternative is to jointly learn the dictionary and classifier by incorporating a classification cost term. However, this will be leading to many parameters to tune, which makes the approach computationally expensive.

.

\FloatBarrier
%\section*{References}

%\bibliography{mybibfile}

\begin{thebibliography}{72}
\expandafter\ifx\csname natexlab\endcsname\relax\def\natexlab#1{#1}\fi
\providecommand{\url}[1]{\texttt{#1}}
\providecommand{\href}[2]{#2}
\providecommand{\path}[1]{#1}
\providecommand{\DOIprefix}{doi:}
\providecommand{\ArXivprefix}{arXiv:}
\providecommand{\URLprefix}{URL: }
\providecommand{\Pubmedprefix}{pmid:}
\providecommand{\doi}[1]{\href{http://dx.doi.org/#1}{\path{#1}}}
\providecommand{\Pubmed}[1]{\href{pmid:#1}{\path{#1}}}
\providecommand{\bibinfo}[2]{#2}
\ifx\xfnm\relax \def\xfnm[#1]{\unskip,\space#1}\fi
%Type = Article
\bibitem[{Aharon et~al.(2006)Aharon, Elad and Bruckstein}]{aharon2006img}
\bibinfo{author}{Aharon, M.}, \bibinfo{author}{Elad, M.},
  \bibinfo{author}{Bruckstein, A.}, \bibinfo{year}{2006}.
\newblock \bibinfo{title}{{$\rm$} k -svd: An algorithm for designing
  overcomplete dictionaries for sparse representation}.
\newblock \bibinfo{journal}{IEEE Transactions on Signal Processing}
  \bibinfo{volume}{54}, \bibinfo{pages}{4311--4322}.
\newblock \DOIprefix\doi{10.1109/TSP.2006.881199}.
%Type = Article
\bibitem[{Al~Maadeed et~al.(2018)Al~Maadeed, Jiang, Rida and
  Bouridane}]{al2018palmprint}
\bibinfo{author}{Al~Maadeed, S.}, \bibinfo{author}{Jiang, X.},
  \bibinfo{author}{Rida, I.}, \bibinfo{author}{Bouridane, A.},
  \bibinfo{year}{2018}.
\newblock \bibinfo{title}{Palmprint identification using sparse and dense
  hybrid representation}.
\newblock \bibinfo{journal}{Multimedia Tools and Applications} ,
  \bibinfo{pages}{1--15}\DOIprefix\doi{10.1007/s11042-018-5655-8}.
%Type = Article
\bibitem[{Aucouturier et~al.(2007)Aucouturier, Defreville and
  Pachet}]{aucouturier2007bag}
\bibinfo{author}{Aucouturier, J.J.}, \bibinfo{author}{Defreville, B.},
  \bibinfo{author}{Pachet, F.}, \bibinfo{year}{2007}.
\newblock \bibinfo{title}{The bag-of-frames approach to audio pattern
  recognition: A sufficient model for urban soundscapes but not for polyphonic
  music}.
\newblock \bibinfo{journal}{The Journal of the Acoustical Society of America}
  \bibinfo{volume}{122}, \bibinfo{pages}{881--891}.
%Type = Article
\bibitem[{Barchiesi et~al.(2015)Barchiesi, Giannoulis, Stowell and
  Plumbley}]{barchiesi2015acoustic}
\bibinfo{author}{Barchiesi, D.}, \bibinfo{author}{Giannoulis, D.},
  \bibinfo{author}{Stowell, D.}, \bibinfo{author}{Plumbley, M.D.},
  \bibinfo{year}{2015}.
\newblock \bibinfo{title}{Acoustic scene classification: Classifying
  environments from the sounds they produce}.
\newblock \bibinfo{journal}{IEEE Signal Processing Magazine}
  \bibinfo{volume}{32}, \bibinfo{pages}{16--34}.
\newblock \DOIprefix\doi{10.1109/MSP.2014.2326181}.
%Type = Inproceedings
\bibitem[{Benetos et~al.(2012)Benetos, Lagrange and
  Dixon}]{benetos2012characterisation}
\bibinfo{author}{Benetos, E.}, \bibinfo{author}{Lagrange, M.},
  \bibinfo{author}{Dixon, S.}, \bibinfo{year}{2012}.
\newblock \bibinfo{title}{Characterisation of acoustic scenes using a
  temporally constrained shift-invariant model}, in: \bibinfo{booktitle}{DAFx}.
%Type = Article
\bibitem[{Benzeghiba et~al.(2007)Benzeghiba, Mori, Deroo, Dupont, Erbes,
  Jouvet, Fissore, Laface, Mertins, Ris, Rose, Tyagi and
  Wellekens}]{Benzeghiba2007763}
\bibinfo{author}{Benzeghiba, M.}, \bibinfo{author}{Mori, R.D.},
  \bibinfo{author}{Deroo, O.}, \bibinfo{author}{Dupont, S.},
  \bibinfo{author}{Erbes, T.}, \bibinfo{author}{Jouvet, D.},
  \bibinfo{author}{Fissore, L.}, \bibinfo{author}{Laface, P.},
  \bibinfo{author}{Mertins, A.}, \bibinfo{author}{Ris, C.},
  \bibinfo{author}{Rose, R.}, \bibinfo{author}{Tyagi, V.},
  \bibinfo{author}{Wellekens, C.}, \bibinfo{year}{2007}.
\newblock \bibinfo{title}{Automatic speech recognition and speech variability:
  A review}.
\newblock \bibinfo{journal}{Speech Communication} \bibinfo{volume}{49},
  \bibinfo{pages}{763 -- 786}.
\newblock \DOIprefix\doi{http://dx.doi.org/10.1016/j.specom.2007.02.006}.
%Type = Book
\bibitem[{Bertsekas(1999)}]{bertsekas1999nonlinear}
\bibinfo{author}{Bertsekas, D.P.}, \bibinfo{year}{1999}.
\newblock \bibinfo{title}{Nonlinear programming}.
\newblock \bibinfo{publisher}{Athena scientific Belmont}.
%Type = Article
\bibitem[{Biem et~al.(2001)Biem, Katagiri, McDermott and
  Juang}]{biem2001application}
\bibinfo{author}{Biem, A.}, \bibinfo{author}{Katagiri, S.},
  \bibinfo{author}{McDermott, E.}, \bibinfo{author}{Juang, B.H.},
  \bibinfo{year}{2001}.
\newblock \bibinfo{title}{An application of discriminative feature extraction
  to filter-bank-based speech recognition}.
\newblock \bibinfo{journal}{IEEE Transactions on Speech and Audio Processing}
  \bibinfo{volume}{9}, \bibinfo{pages}{96--110}.
\newblock \DOIprefix\doi{10.1109/89.902277}.
%Type = Article
\bibitem[{Cauchi(2011)}]{cauchi2011}
\bibinfo{author}{Cauchi, B.}, \bibinfo{year}{2011}.
\newblock \bibinfo{title}{Non-negative matrix factorisation applied to auditory
  scenes classification}.
\newblock \bibinfo{journal}{Master's thesis, Master ATIAM, Universit{\'e}
  Pierre et Marie Curie} .
%Type = Inproceedings
\bibitem[{Cotton and Ellis(2011)}]{cotton2011spectral}
\bibinfo{author}{Cotton, C.V.}, \bibinfo{author}{Ellis, D.P.W.},
  \bibinfo{year}{2011}.
\newblock \bibinfo{title}{Spectral vs. spectro-temporal features for acoustic
  event detection}, in: \bibinfo{booktitle}{2011 IEEE Workshop on Applications
  of Signal Processing to Audio and Acoustics (WASPAA)}, pp.
  \bibinfo{pages}{69--72}.
\newblock \DOIprefix\doi{10.1109/ASPAA.2011.6082331}.
%Type = Article
\bibitem[{Davis and Mermelstein(1980)}]{davis1980comparison}
\bibinfo{author}{Davis, S.}, \bibinfo{author}{Mermelstein, P.},
  \bibinfo{year}{1980}.
\newblock \bibinfo{title}{Comparison of parametric representations for
  monosyllabic word recognition in continuously spoken sentences}.
\newblock \bibinfo{journal}{IEEE Transactions on Acoustics, Speech, and Signal
  Processing} \bibinfo{volume}{28}, \bibinfo{pages}{357--366}.
\newblock \DOIprefix\doi{10.1109/TASSP.1980.1163420}.
%Type = Article
\bibitem[{Davy et~al.(2002)Davy, Gretton, Doucet and
  Rayner}]{davy2002optimized}
\bibinfo{author}{Davy, M.}, \bibinfo{author}{Gretton, A.},
  \bibinfo{author}{Doucet, A.}, \bibinfo{author}{Rayner, P.J.W.},
  \bibinfo{year}{2002}.
\newblock \bibinfo{title}{Optimized support vector machines for nonstationary
  signal classification}.
\newblock \bibinfo{journal}{IEEE Signal Processing Letters}
  \bibinfo{volume}{9}, \bibinfo{pages}{442--445}.
\newblock \DOIprefix\doi{10.1109/LSP.2002.806070}.
%Type = Article
\bibitem[{Dennis et~al.(2013)Dennis, Tran and Chng}]{dennis2013image}
\bibinfo{author}{Dennis, J.}, \bibinfo{author}{Tran, H.D.},
  \bibinfo{author}{Chng, E.S.}, \bibinfo{year}{2013}.
\newblock \bibinfo{title}{Image feature representation of the subband power
  distribution for robust sound event classification}.
\newblock \bibinfo{journal}{IEEE Transactions on Audio, Speech, and Language
  Processing} \bibinfo{volume}{21}, \bibinfo{pages}{367--377}.
\newblock \DOIprefix\doi{10.1109/TASL.2012.2226160}.
%Type = Article
\bibitem[{Eksioglu(2014)}]{eksioglu2014online}
\bibinfo{author}{Eksioglu, E.M.}, \bibinfo{year}{2014}.
\newblock \bibinfo{title}{Online dictionary learning algorithm with periodic
  updates and its application to image denoising}.
\newblock \bibinfo{journal}{Expert Systems with Applications}
  \bibinfo{volume}{41}, \bibinfo{pages}{3682--3690}.
%Type = Article
\bibitem[{Elad and Aharon(2006)}]{elad2006}
\bibinfo{author}{Elad, M.}, \bibinfo{author}{Aharon, M.}, \bibinfo{year}{2006}.
\newblock \bibinfo{title}{Image denoising via sparse and redundant
  representations over learned dictionaries}.
\newblock \bibinfo{journal}{IEEE Transactions on Image Processing}
  \bibinfo{volume}{15}, \bibinfo{pages}{3736--3745}.
\newblock \DOIprefix\doi{10.1109/TIP.2006.881969}.
%Type = Article
\bibitem[{Elad et~al.(2010)Elad, Figueiredo and Ma}]{elad2010}
\bibinfo{author}{Elad, M.}, \bibinfo{author}{Figueiredo, M.A.T.},
  \bibinfo{author}{Ma, Y.}, \bibinfo{year}{2010}.
\newblock \bibinfo{title}{On the role of sparse and redundant representations
  in image processing}.
\newblock \bibinfo{journal}{Proceedings of the IEEE} \bibinfo{volume}{98},
  \bibinfo{pages}{972--982}.
\newblock \DOIprefix\doi{10.1109/JPROC.2009.2037655}.
%Type = Article
\bibitem[{Flores and Scharcanski(2016)}]{flores2016segmentation}
\bibinfo{author}{Flores, E.}, \bibinfo{author}{Scharcanski, J.},
  \bibinfo{year}{2016}.
\newblock \bibinfo{title}{Segmentation of melanocytic skin lesions using
  feature learning and dictionaries}.
\newblock \bibinfo{journal}{Expert Systems with Applications}
  \bibinfo{volume}{56}, \bibinfo{pages}{300--309}.
%Type = Inproceedings
\bibitem[{Fujishima(1999)}]{fujichroma}
\bibinfo{author}{Fujishima, T.}, \bibinfo{year}{1999}.
\newblock \bibinfo{title}{Realtime chord recognition of musical sound: a system
  using common lisp music.}, in: \bibinfo{booktitle}{ICMC}, pp.
  \bibinfo{pages}{464--467}.
%Type = Incollection
\bibitem[{Fulkerson et~al.(2008)Fulkerson, Vedaldi and
  Soatto}]{fulkerson2008localizing}
\bibinfo{author}{Fulkerson, B.}, \bibinfo{author}{Vedaldi, A.},
  \bibinfo{author}{Soatto, S.}, \bibinfo{year}{2008}.
\newblock \bibinfo{title}{Localizing objects with smart dictionaries}, in:
  \bibinfo{editor}{Forsyth, D.}, \bibinfo{editor}{Torr, P.},
  \bibinfo{editor}{Zisserman, A.} (Eds.), \bibinfo{booktitle}{Computer Vision
  -- ECCV 2008: 10th European Conference on Computer Vision, Marseille, France,
  October 12-18, 2008, Proceedings, Part I}. \bibinfo{publisher}{Springer
  Berlin Heidelberg}, pp. \bibinfo{pages}{179--192}.
\newblock \DOIprefix\doi{10.1007/978-3-540-88682-2_15}.
%Type = Article
\bibitem[{Gangeh et~al.(2015)Gangeh, Farahat, Ghodsi and
  Kamel}]{gangeh2015supervised}
\bibinfo{author}{Gangeh, M.J.}, \bibinfo{author}{Farahat, A.K.},
  \bibinfo{author}{Ghodsi, A.}, \bibinfo{author}{Kamel, M.S.},
  \bibinfo{year}{2015}.
\newblock \bibinfo{title}{Supervised dictionary learning and sparse
  representation-a review}.
\newblock \bibinfo{journal}{arXiv preprint arXiv:1502.05928} .
%Type = Inproceedings
\bibitem[{Geiger et~al.(2013)Geiger, Schuller and Rigoll}]{geiger2013large}
\bibinfo{author}{Geiger, J.T.}, \bibinfo{author}{Schuller, B.},
  \bibinfo{author}{Rigoll, G.}, \bibinfo{year}{2013}.
\newblock \bibinfo{title}{Large-scale audio feature extraction and svm for
  acoustic scene classification}, in: \bibinfo{booktitle}{2013 IEEE Workshop on
  Applications of Signal Processing to Audio and Acoustics}, pp.
  \bibinfo{pages}{1--4}.
\newblock \DOIprefix\doi{10.1109/WASPAA.2013.6701857}.
%Type = Inproceedings
\bibitem[{Honeine et~al.(2006)Honeine, Richard, Flandrin and
  Pothin}]{honeine2006optimal}
\bibinfo{author}{Honeine, P.}, \bibinfo{author}{Richard, C.},
  \bibinfo{author}{Flandrin, P.}, \bibinfo{author}{Pothin, J.B.},
  \bibinfo{year}{2006}.
\newblock \bibinfo{title}{Optimal selection of time-frequency representations
  for signal classification: a kernel-target alignment approach}, in:
  \bibinfo{booktitle}{2006 IEEE International Conference on Acoustics Speech
  and Signal Processing Proceedings}, pp. \bibinfo{pages}{III--III}.
\newblock \DOIprefix\doi{10.1109/ICASSP.2006.1660694}.
%Type = Inproceedings
\bibitem[{Hu et~al.(2012)Hu, Liu, Jiang et~al.}]{hu2012combining}
\bibinfo{author}{Hu, P.}, \bibinfo{author}{Liu, W.}, \bibinfo{author}{Jiang,
  W.}, et~al., \bibinfo{year}{2012}.
\newblock \bibinfo{title}{Combining frame and segment based models for
  environmental sound classification.}, in: \bibinfo{booktitle}{INTERSPEECH},
  pp. \bibinfo{pages}{2502--2505}.
%Type = Article
\bibitem[{Kinnunen et~al.(2012)Kinnunen, Saeidi, Sedlak, Lee, Sandberg,
  Hansson-Sandsten and Li}]{kinnunen2012low}
\bibinfo{author}{Kinnunen, T.}, \bibinfo{author}{Saeidi, R.},
  \bibinfo{author}{Sedlak, F.}, \bibinfo{author}{Lee, K.A.},
  \bibinfo{author}{Sandberg, J.}, \bibinfo{author}{Hansson-Sandsten, M.},
  \bibinfo{author}{Li, H.}, \bibinfo{year}{2012}.
\newblock \bibinfo{title}{Low-variance multitaper mfcc features: A case study
  in robust speaker verification}.
\newblock \bibinfo{journal}{IEEE Transactions on Audio, Speech, and Language
  Processing} \bibinfo{volume}{20}, \bibinfo{pages}{1990--2001}.
\newblock \DOIprefix\doi{10.1109/TASL.2012.2191960}.
%Type = Inproceedings
\bibitem[{Kronvall et~al.(2015)Kronvall, Juhlin, Adalbjornsson and
  Jakobsson}]{sparsechroma}
\bibinfo{author}{Kronvall, T.}, \bibinfo{author}{Juhlin, M.},
  \bibinfo{author}{Adalbjornsson, S.I.}, \bibinfo{author}{Jakobsson, A.},
  \bibinfo{year}{2015}.
\newblock \bibinfo{title}{Sparse chroma estimation for harmonic audio}, in:
  \bibinfo{booktitle}{2015 IEEE International Conference on Acoustics, Speech
  and Signal Processing (ICASSP)}, pp. \bibinfo{pages}{579--583}.
\newblock \DOIprefix\doi{10.1109/ICASSP.2015.7178035}.
%Type = Article
\bibitem[{Kronvall et~al.(2017)Kronvall, Juhlin, Swärd, Adalbjörnsson and
  Jakobsson}]{Kronvall2017105}
\bibinfo{author}{Kronvall, T.}, \bibinfo{author}{Juhlin, M.},
  \bibinfo{author}{Swärd, J.}, \bibinfo{author}{Adalbjörnsson, S.I.},
  \bibinfo{author}{Jakobsson, A.}, \bibinfo{year}{2017}.
\newblock \bibinfo{title}{Sparse modeling of chroma features}.
\newblock \bibinfo{journal}{Signal Processing} \bibinfo{volume}{130},
  \bibinfo{pages}{105 -- 117}.
\newblock \DOIprefix\doi{http://dx.doi.org/10.1016/j.sigpro.2016.06.020}.
%Type = Article
\bibitem[{Lazebnik and Raginsky(2009)}]{lazebnik2009supervised}
\bibinfo{author}{Lazebnik, S.}, \bibinfo{author}{Raginsky, M.},
  \bibinfo{year}{2009}.
\newblock \bibinfo{title}{Supervised learning of quantizer codebooks by
  information loss minimization}.
\newblock \bibinfo{journal}{IEEE Transactions on Pattern Analysis and Machine
  Intelligence} \bibinfo{volume}{31}, \bibinfo{pages}{1294--1309}.
\newblock \DOIprefix\doi{10.1109/TPAMI.2008.138}.
%Type = Inproceedings
\bibitem[{Lee et~al.(2006)Lee, Battle, Raina and Ng}]{lee2006efficient}
\bibinfo{author}{Lee, H.}, \bibinfo{author}{Battle, A.},
  \bibinfo{author}{Raina, R.}, \bibinfo{author}{Ng, A.Y.},
  \bibinfo{year}{2006}.
\newblock \bibinfo{title}{Efficient sparse coding algorithms}, in:
  \bibinfo{booktitle}{Advances in neural information processing systems}, pp.
  \bibinfo{pages}{801--808}.
%Type = Inproceedings
\bibitem[{Lee(2006)}]{lee2006automatic}
\bibinfo{author}{Lee, K.}, \bibinfo{year}{2006}.
\newblock \bibinfo{title}{Automatic chord recognition from audio using enhanced
  pitch class profile}, in: \bibinfo{booktitle}{Proc. of the International
  Computer Music Conference}, p.~\bibinfo{pages}{26}.
%Type = Inproceedings
\bibitem[{Lee et~al.(2013)Lee, Hyung and Nam}]{lee2013acoustic}
\bibinfo{author}{Lee, K.}, \bibinfo{author}{Hyung, Z.}, \bibinfo{author}{Nam,
  J.}, \bibinfo{year}{2013}.
\newblock \bibinfo{title}{Acoustic scene classification using sparse feature
  learning and event-based pooling}, in: \bibinfo{booktitle}{2013 IEEE Workshop
  on Applications of Signal Processing to Audio and Acoustics}, pp.
  \bibinfo{pages}{1--4}.
\newblock \DOIprefix\doi{10.1109/WASPAA.2013.6701893}.
%Type = Article
\bibitem[{Li et~al.(2014)Li, Deng, Gong and Haeb-Umbach}]{li2014overview}
\bibinfo{author}{Li, J.}, \bibinfo{author}{Deng, L.}, \bibinfo{author}{Gong,
  Y.}, \bibinfo{author}{Haeb-Umbach, R.}, \bibinfo{year}{2014}.
\newblock \bibinfo{title}{An overview of noise-robust automatic speech
  recognition}.
\newblock \bibinfo{journal}{IEEE/ACM Transactions on Audio, Speech, and
  Language Processing} \bibinfo{volume}{22}, \bibinfo{pages}{745--777}.
\newblock \DOIprefix\doi{10.1109/TASLP.2014.2304637}.
%Type = Inproceedings
\bibitem[{Lian et~al.(2010)Lian, Li, Wang, Lu and
  Zhang}]{lian2010probabilistic}
\bibinfo{author}{Lian, X.C.}, \bibinfo{author}{Li, Z.}, \bibinfo{author}{Wang,
  C.}, \bibinfo{author}{Lu, B.L.}, \bibinfo{author}{Zhang, L.},
  \bibinfo{year}{2010}.
\newblock \bibinfo{title}{Probabilistic models for supervised dictionary
  learning}, in: \bibinfo{booktitle}{2010 IEEE Computer Society Conference on
  Computer Vision and Pattern Recognition}, pp. \bibinfo{pages}{2305--2312}.
\newblock \DOIprefix\doi{10.1109/CVPR.2010.5539915}.
%Type = Article
\bibitem[{Lyon(2010)}]{lyon2010machine}
\bibinfo{author}{Lyon, R.F.}, \bibinfo{year}{2010}.
\newblock \bibinfo{title}{Machine hearing: An emerging field [exploratory
  dsp]}.
\newblock \bibinfo{journal}{IEEE Signal Processing Magazine}
  \bibinfo{volume}{27}, \bibinfo{pages}{131--139}.
\newblock \DOIprefix\doi{10.1109/MSP.2010.937498}.
%Type = Article
\bibitem[{Ma et~al.(2006)Ma, Milner and Smith}]{ma2006acoustic}
\bibinfo{author}{Ma, L.}, \bibinfo{author}{Milner, B.}, \bibinfo{author}{Smith,
  D.}, \bibinfo{year}{2006}.
\newblock \bibinfo{title}{Acoustic environment classification}.
\newblock \bibinfo{journal}{ACM Trans. Speech Lang. Process.}
  \bibinfo{volume}{3}, \bibinfo{pages}{1--22}.
\newblock \DOIprefix\doi{10.1145/1149290.1149292}.
%Type = Inproceedings
\bibitem[{Ma et~al.(2003)Ma, Smith and Milner}]{ma2003context}
\bibinfo{author}{Ma, L.}, \bibinfo{author}{Smith, D.}, \bibinfo{author}{Milner,
  B.P.}, \bibinfo{year}{2003}.
\newblock \bibinfo{title}{Context awareness using environmental noise
  classification.}, in: \bibinfo{booktitle}{INTERSPEECH}, pp.
  \bibinfo{pages}{2237--2240}.
%Type = Article
\bibitem[{Mairal et~al.(2012)Mairal, Bach and Ponce}]{mairal2012task}
\bibinfo{author}{Mairal, J.}, \bibinfo{author}{Bach, F.},
  \bibinfo{author}{Ponce, J.}, \bibinfo{year}{2012}.
\newblock \bibinfo{title}{Task-driven dictionary learning}.
\newblock \bibinfo{journal}{IEEE Transactions on Pattern Analysis and Machine
  Intelligence} \bibinfo{volume}{34}, \bibinfo{pages}{791--804}.
\newblock \DOIprefix\doi{10.1109/TPAMI.2011.156}.
%Type = Article
\bibitem[{Mairal et~al.(2008)Mairal, Elad and Sapiro}]{mairal2008}
\bibinfo{author}{Mairal, J.}, \bibinfo{author}{Elad, M.},
  \bibinfo{author}{Sapiro, G.}, \bibinfo{year}{2008}.
\newblock \bibinfo{title}{Sparse representation for color image restoration}.
\newblock \bibinfo{journal}{IEEE Transactions on Image Processing}
  \bibinfo{volume}{17}, \bibinfo{pages}{53--69}.
\newblock \DOIprefix\doi{10.1109/TIP.2007.911828}.
%Type = Inproceedings
\bibitem[{Mairal et~al.(2009)Mairal, Ponce, Sapiro, Zisserman and
  Bach}]{mairal2009}
\bibinfo{author}{Mairal, J.}, \bibinfo{author}{Ponce, J.},
  \bibinfo{author}{Sapiro, G.}, \bibinfo{author}{Zisserman, A.},
  \bibinfo{author}{Bach, F.R.}, \bibinfo{year}{2009}.
\newblock \bibinfo{title}{Supervised dictionary learning}, in:
  \bibinfo{booktitle}{Advances in neural information processing systems}, pp.
  \bibinfo{pages}{1033--1040}.
%Type = Inproceedings
\bibitem[{Oudre et~al.(2009)Oudre, Grenier and F{\'e}votte}]{oudre2009template}
\bibinfo{author}{Oudre, L.}, \bibinfo{author}{Grenier, Y.},
  \bibinfo{author}{F{\'e}votte, C.}, \bibinfo{year}{2009}.
\newblock \bibinfo{title}{Template-based chord recognition: Influence of the
  chord types.}, in: \bibinfo{booktitle}{ISMIR}, pp. \bibinfo{pages}{153--158}.
%Type = Article
\bibitem[{Oudre et~al.(2011)Oudre, Grenier and Fevotte}]{oudre2011chord}
\bibinfo{author}{Oudre, L.}, \bibinfo{author}{Grenier, Y.},
  \bibinfo{author}{Fevotte, C.}, \bibinfo{year}{2011}.
\newblock \bibinfo{title}{Chord recognition by fitting rescaled chroma vectors
  to chord templates}.
\newblock \bibinfo{journal}{IEEE Transactions on Audio, Speech, and Language
  Processing} \bibinfo{volume}{19}, \bibinfo{pages}{2222--2233}.
\newblock \DOIprefix\doi{10.1109/TASL.2011.2139205}.
%Type = Inproceedings
\bibitem[{Papadopoulos and Peeters(2008)}]{papadopoulos2008simultaneous}
\bibinfo{author}{Papadopoulos, H.}, \bibinfo{author}{Peeters, G.},
  \bibinfo{year}{2008}.
\newblock \bibinfo{title}{Simultaneous estimation of chord progression and
  downbeats from an audio file}, in: \bibinfo{booktitle}{2008 IEEE
  International Conference on Acoustics, Speech and Signal Processing}, pp.
  \bibinfo{pages}{121--124}.
\newblock \DOIprefix\doi{10.1109/ICASSP.2008.4517561}.
%Type = Article
\bibitem[{Petersen et~al.(2008)Petersen, Pedersen et~al.}]{petersen2008matrix}
\bibinfo{author}{Petersen, K.B.}, \bibinfo{author}{Pedersen, M.S.}, et~al.,
  \bibinfo{year}{2008}.
\newblock \bibinfo{title}{The matrix cookbook}.
\newblock \bibinfo{journal}{Technical University of Denmark}
  \bibinfo{volume}{7}, \bibinfo{pages}{15}.
%Type = Article
\bibitem[{Phaisangittisagul et~al.(2017)Phaisangittisagul, Thainimit and
  Chen}]{phaisangittisagul2017predictive}
\bibinfo{author}{Phaisangittisagul, E.}, \bibinfo{author}{Thainimit, S.},
  \bibinfo{author}{Chen, W.}, \bibinfo{year}{2017}.
\newblock \bibinfo{title}{Predictive high-level feature representation based on
  dictionary learning}.
\newblock \bibinfo{journal}{Expert Systems with Applications}
  \bibinfo{volume}{69}, \bibinfo{pages}{101--109}.
%Type = Article
\bibitem[{Rakotomamonjy and Gasso(2015)}]{Rakotomamonjy:2015}
\bibinfo{author}{Rakotomamonjy, A.}, \bibinfo{author}{Gasso, G.},
  \bibinfo{year}{2015}.
\newblock \bibinfo{title}{Histogram of gradients of time-frequency
  representations for audio scene classification}.
\newblock \bibinfo{journal}{IEEE/ACM Trans. Audio, Speech and Lang. Proc.}
  \bibinfo{volume}{23}, \bibinfo{pages}{142--153}.
\newblock \DOIprefix\doi{10.1109/TASLP.2014.2375575}.
%Type = Inproceedings
\bibitem[{Ramirez et~al.(2010)Ramirez, Sprechmann and Sapiro}]{ramirez2010}
\bibinfo{author}{Ramirez, I.}, \bibinfo{author}{Sprechmann, P.},
  \bibinfo{author}{Sapiro, G.}, \bibinfo{year}{2010}.
\newblock \bibinfo{title}{Classification and clustering via dictionary learning
  with structured incoherence and shared features}, in:
  \bibinfo{booktitle}{2010 IEEE Computer Society Conference on Computer Vision
  and Pattern Recognition}, pp. \bibinfo{pages}{3501--3508}.
\newblock \DOIprefix\doi{10.1109/CVPR.2010.5539964}.
%Type = Article
\bibitem[{Rida(2018)}]{rida2018feature}
\bibinfo{author}{Rida, I.}, \bibinfo{year}{2018}.
\newblock \bibinfo{title}{Feature extraction for temporal signal recognition:
  An overview}.
\newblock \bibinfo{journal}{arXiv preprint arXiv:1812.01780} .
%Type = Inproceedings
\bibitem[{Rida et~al.(2018a)Rida, Al~Maadeed and Al~Maadeed}]{rida2018novel}
\bibinfo{author}{Rida, I.}, \bibinfo{author}{Al~Maadeed, N.},
  \bibinfo{author}{Al~Maadeed, S.}, \bibinfo{year}{2018}a.
\newblock \bibinfo{title}{A novel efficient classwise sparse and collaborative
  representation for holistic palmprint recognition}, in:
  \bibinfo{booktitle}{2018 NASA/ESA Conference on Adaptive Hardware and Systems
  (AHS)}, \bibinfo{organization}{IEEE}. pp. \bibinfo{pages}{156--161}.
\newblock \DOIprefix\doi{10.1109/AHS.2018.8541428}.
%Type = Article
\bibitem[{Rida et~al.(2018b)Rida, Al-Maadeed, Al-Maadeed and
  Bakshi}]{rida2018comprehensive}
\bibinfo{author}{Rida, I.}, \bibinfo{author}{Al-Maadeed, N.},
  \bibinfo{author}{Al-Maadeed, S.}, \bibinfo{author}{Bakshi, S.},
  \bibinfo{year}{2018}b.
\newblock \bibinfo{title}{A comprehensive overview of feature representation
  for biometric recognition}.
\newblock \bibinfo{journal}{Multimedia Tools and Applications} ,
  \bibinfo{pages}{1--24}\DOIprefix\doi{10.1007/s11042-018-6808-5}.
%Type = Inproceedings
\bibitem[{Rida et~al.(2015a)Rida, Al~Maadeed and
  Bouridane}]{rida2015unsupervised}
\bibinfo{author}{Rida, I.}, \bibinfo{author}{Al~Maadeed, S.},
  \bibinfo{author}{Bouridane, A.}, \bibinfo{year}{2015}a.
\newblock \bibinfo{title}{Unsupervised feature selection method for improved
  human gait recognition}, in: \bibinfo{booktitle}{Signal Processing Conference
  (EUSIPCO), 2015 23rd European}, \bibinfo{organization}{IEEE}. pp.
  \bibinfo{pages}{1128--1132}.
%Type = Inproceedings
\bibitem[{Rida et~al.(2018c)Rida, Al~Maadeed, Jiang, Lunke and
  Bensrhair}]{rida2018ensemble}
\bibinfo{author}{Rida, I.}, \bibinfo{author}{Al~Maadeed, S.},
  \bibinfo{author}{Jiang, X.}, \bibinfo{author}{Lunke, F.},
  \bibinfo{author}{Bensrhair, A.}, \bibinfo{year}{2018}c.
\newblock \bibinfo{title}{An ensemble learning method based on random subspace
  sampling for palmprint identification}, in: \bibinfo{booktitle}{2018 IEEE
  International conference on acoustics, speech and signal processing
  (ICASSP)}, \bibinfo{organization}{IEEE}. pp. \bibinfo{pages}{2047--2051}.
%Type = Article
\bibitem[{Rida et~al.(2018d)Rida, Al-Maadeed, Mahmood, Bouridane and
  Bakshi}]{8244323}
\bibinfo{author}{Rida, I.}, \bibinfo{author}{Al-Maadeed, S.},
  \bibinfo{author}{Mahmood, A.}, \bibinfo{author}{Bouridane, A.},
  \bibinfo{author}{Bakshi, S.}, \bibinfo{year}{2018}d.
\newblock \bibinfo{title}{Palmprint identification using an ensemble of sparse
  representations}.
\newblock \bibinfo{journal}{IEEE Access} \bibinfo{volume}{6},
  \bibinfo{pages}{3241--3248}.
\newblock \DOIprefix\doi{10.1109/ACCESS.2017.2787666}.
%Type = Article
\bibitem[{Rida et~al.(2016a)Rida, Almaadeed and Bouridane}]{rida2016gaitt}
\bibinfo{author}{Rida, I.}, \bibinfo{author}{Almaadeed, S.},
  \bibinfo{author}{Bouridane, A.}, \bibinfo{year}{2016}a.
\newblock \bibinfo{title}{Gait recognition based on modified phase-only
  correlation}.
\newblock \bibinfo{journal}{Signal, Image and Video Processing}
  \bibinfo{volume}{10}, \bibinfo{pages}{463--470}.
\newblock \DOIprefix\doi{10.1007/s11760-015-0766-4}.
%Type = Inproceedings
\bibitem[{Rida et~al.(2015b)Rida, Bouridane, Marcialis and
  Tuveri}]{rida2015improved}
\bibinfo{author}{Rida, I.}, \bibinfo{author}{Bouridane, A.},
  \bibinfo{author}{Marcialis, G.L.}, \bibinfo{author}{Tuveri, P.},
  \bibinfo{year}{2015}b.
\newblock \bibinfo{title}{Improved human gait recognition}, in:
  \bibinfo{booktitle}{International Conference on Image Analysis and
  Processing}, \bibinfo{organization}{Springer}. pp. \bibinfo{pages}{119--129}.
\newblock \DOIprefix\doi{10.1007/978-3-319-23234-8\_12}.
%Type = Inproceedings
\bibitem[{Rida et~al.(2014)Rida, Herault and Gasso}]{rida2014supervised}
\bibinfo{author}{Rida, I.}, \bibinfo{author}{Herault, R.},
  \bibinfo{author}{Gasso, G.}, \bibinfo{year}{2014}.
\newblock \bibinfo{title}{Supervised music chord recognition}, in:
  \bibinfo{booktitle}{2014 13th International Conference on Machine Learning
  and Applications}, pp. \bibinfo{pages}{336--341}.
\newblock \DOIprefix\doi{10.1109/ICMLA.2014.60}.
%Type = Article
\bibitem[{Rida et~al.(2018e)Rida, Herault, Marcialis and
  Gasso}]{rida2018palmprintt}
\bibinfo{author}{Rida, I.}, \bibinfo{author}{Herault, R.},
  \bibinfo{author}{Marcialis, G.L.}, \bibinfo{author}{Gasso, G.},
  \bibinfo{year}{2018}e.
\newblock \bibinfo{title}{Palmprint recognition with an efficient data driven
  ensemble classifier}.
\newblock \bibinfo{journal}{Pattern Recognition Letters}
  \DOIprefix\doi{10.1016/j.patrec.2018.04.033}.
%Type = Article
\bibitem[{Rida et~al.(2016b)Rida, Jiang and Marcialis}]{7350221}
\bibinfo{author}{Rida, I.}, \bibinfo{author}{Jiang, X.},
  \bibinfo{author}{Marcialis, G.L.}, \bibinfo{year}{2016}b.
\newblock \bibinfo{title}{Human body part selection by group lasso of motion
  for model-free gait recognition}.
\newblock \bibinfo{journal}{IEEE Signal Processing Letters}
  \bibinfo{volume}{23}, \bibinfo{pages}{154--158}.
\newblock \DOIprefix\doi{10.1109/LSP.2015.2507200}.
%Type = Inproceedings
\bibitem[{Roma et~al.(2013)Roma, Nogueira and Herrera}]{roma2013recurrence}
\bibinfo{author}{Roma, G.}, \bibinfo{author}{Nogueira, W.},
  \bibinfo{author}{Herrera, P.}, \bibinfo{year}{2013}.
\newblock \bibinfo{title}{Recurrence quantification analysis features for
  environmental sound recognition}, in: \bibinfo{booktitle}{2013 IEEE Workshop
  on Applications of Signal Processing to Audio and Acoustics}, pp.
  \bibinfo{pages}{1--4}.
\newblock \DOIprefix\doi{10.1109/WASPAA.2013.6701890}.
%Type = Article
\bibitem[{Sangnier et~al.(2015)Sangnier, Gauthier and
  Rakotomamonjy}]{sangnier2015filter}
\bibinfo{author}{Sangnier, M.}, \bibinfo{author}{Gauthier, J.},
  \bibinfo{author}{Rakotomamonjy, A.}, \bibinfo{year}{2015}.
\newblock \bibinfo{title}{Filter bank learning for signal classification}.
\newblock \bibinfo{journal}{Signal Processing} \bibinfo{volume}{113},
  \bibinfo{pages}{124 -- 137}.
\newblock \DOIprefix\doi{http://dx.doi.org/10.1016/j.sigpro.2014.12.028}.
%Type = Book
\bibitem[{Sch{\"o}lkopf and Smola(2002)}]{scholkopf2002learning}
\bibinfo{author}{Sch{\"o}lkopf, B.}, \bibinfo{author}{Smola, A.J.},
  \bibinfo{year}{2002}.
\newblock \bibinfo{title}{Learning with kernels: support vector machines,
  regularization, optimization, and beyond}.
\newblock \bibinfo{publisher}{MIT press}.
%Type = Inproceedings
\bibitem[{Sheh and Ellis(2003)}]{sheh2003chord}
\bibinfo{author}{Sheh, A.}, \bibinfo{author}{Ellis, D.P.},
  \bibinfo{year}{2003}.
\newblock \bibinfo{title}{Chord segmentation and recognition using em-trained
  hidden markov models.}, in: \bibinfo{booktitle}{ISMIR}, pp.
  \bibinfo{pages}{183--189}.
%Type = Article
\bibitem[{Strauss et~al.(2003)Strauss, Steidl and Delb}]{strauss2003feature}
\bibinfo{author}{Strauss, D.J.}, \bibinfo{author}{Steidl, G.},
  \bibinfo{author}{Delb, W.}, \bibinfo{year}{2003}.
\newblock \bibinfo{title}{Feature extraction by shape-adapted local
  discriminant bases}.
\newblock \bibinfo{journal}{Signal Processing} \bibinfo{volume}{83},
  \bibinfo{pages}{359 -- 376}.
\newblock \DOIprefix\doi{http://dx.doi.org/10.1016/S0165-1684(02)00420-6}.
%Type = Article
\bibitem[{T{\"u}ys{\"u}zo{\u{g}}lu and Yaslan(2018)}]{tuysuzouglu2018sparse}
\bibinfo{author}{T{\"u}ys{\"u}zo{\u{g}}lu, G.}, \bibinfo{author}{Yaslan, Y.},
  \bibinfo{year}{2018}.
\newblock \bibinfo{title}{Sparse coding based classifier ensembles in
  supervised and active learning scenarios for data classification}.
\newblock \bibinfo{journal}{Expert Systems with Applications}
  \bibinfo{volume}{91}, \bibinfo{pages}{364--373}.
%Type = Article
\bibitem[{Varma and Zisserman(2009)}]{varma2009statistical}
\bibinfo{author}{Varma, M.}, \bibinfo{author}{Zisserman, A.},
  \bibinfo{year}{2009}.
\newblock \bibinfo{title}{A statistical approach to material classification
  using image patch exemplars}.
\newblock \bibinfo{journal}{IEEE Transactions on Pattern Analysis and Machine
  Intelligence} \bibinfo{volume}{31}, \bibinfo{pages}{2032--2047}.
\newblock \DOIprefix\doi{10.1109/TPAMI.2008.182}.
%Type = Inproceedings
\bibitem[{Weller et~al.(2009)Weller, Ellis and Jebara}]{weller2009structured}
\bibinfo{author}{Weller, A.}, \bibinfo{author}{Ellis, D.},
  \bibinfo{author}{Jebara, T.}, \bibinfo{year}{2009}.
\newblock \bibinfo{title}{Structured prediction models for chord transcription
  of music audio}, in: \bibinfo{booktitle}{2009 International Conference on
  Machine Learning and Applications}, pp. \bibinfo{pages}{590--595}.
\newblock \DOIprefix\doi{10.1109/ICMLA.2009.132}.
%Type = Inproceedings
\bibitem[{Winn et~al.(2005)Winn, Criminisi and Minka}]{winn2005object}
\bibinfo{author}{Winn, J.}, \bibinfo{author}{Criminisi, A.},
  \bibinfo{author}{Minka, T.}, \bibinfo{year}{2005}.
\newblock \bibinfo{title}{Object categorization by learned universal visual
  dictionary}, in: \bibinfo{booktitle}{Tenth IEEE International Conference on
  Computer Vision (ICCV'05) Volume 1}, pp. \bibinfo{pages}{1800--1807 Vol. 2}.
\newblock \DOIprefix\doi{10.1109/ICCV.2005.171}.
%Type = Inproceedings
\bibitem[{Yang et~al.(2011)Yang, Zhang, Feng and Zhang}]{yang2011}
\bibinfo{author}{Yang, M.}, \bibinfo{author}{Zhang, L.}, \bibinfo{author}{Feng,
  X.}, \bibinfo{author}{Zhang, D.}, \bibinfo{year}{2011}.
\newblock \bibinfo{title}{Fisher discrimination dictionary learning for sparse
  representation}, in: \bibinfo{booktitle}{2011 International Conference on
  Computer Vision}, pp. \bibinfo{pages}{543--550}.
\newblock \DOIprefix\doi{10.1109/ICCV.2011.6126286}.
%Type = Inproceedings
\bibitem[{Yang et~al.(2010)Yang, Zhang, Yang and Zhang}]{yang2010}
\bibinfo{author}{Yang, M.}, \bibinfo{author}{Zhang, L.}, \bibinfo{author}{Yang,
  J.}, \bibinfo{author}{Zhang, D.}, \bibinfo{year}{2010}.
\newblock \bibinfo{title}{Metaface learning for sparse representation based
  face recognition}, in: \bibinfo{booktitle}{2010 IEEE International Conference
  on Image Processing}, pp. \bibinfo{pages}{1601--1604}.
\newblock \DOIprefix\doi{10.1109/ICIP.2010.5652363}.
%Type = Article
\bibitem[{Yger and Rakotomamonjy(2011)}]{yger2011wavelet}
\bibinfo{author}{Yger, F.}, \bibinfo{author}{Rakotomamonjy, A.},
  \bibinfo{year}{2011}.
\newblock \bibinfo{title}{Wavelet kernel learning}.
\newblock \bibinfo{journal}{Pattern Recognition} \bibinfo{volume}{44},
  \bibinfo{pages}{2614 -- 2629}.
\newblock \DOIprefix\doi{http://dx.doi.org/10.1016/j.patcog.2011.03.006}.
%Type = Inproceedings
\bibitem[{Yu and Slotine(2009)}]{yu2008audio}
\bibinfo{author}{Yu, G.}, \bibinfo{author}{Slotine, J.J.},
  \bibinfo{year}{2009}.
\newblock \bibinfo{title}{Audio classification from time-frequency texture},
  in: \bibinfo{booktitle}{Acoustics, Speech and Signal Processing, 2009. ICASSP
  2009. IEEE International Conference on}, \bibinfo{organization}{IEEE}. pp.
  \bibinfo{pages}{1677--1680}.
%Type = Article
\bibitem[{Zhang et~al.(2013)Zhang, Zhang and Huang}]{zhang2013simultaneous}
\bibinfo{author}{Zhang, H.}, \bibinfo{author}{Zhang, Y.},
  \bibinfo{author}{Huang, T.S.}, \bibinfo{year}{2013}.
\newblock \bibinfo{title}{Simultaneous discriminative projection and dictionary
  learning for sparse representation based classification}.
\newblock \bibinfo{journal}{Pattern Recognition} \bibinfo{volume}{46},
  \bibinfo{pages}{346 -- 354}.
\newblock \DOIprefix\doi{http://dx.doi.org/10.1016/j.patcog.2012.07.010}.
%Type = Inproceedings
\bibitem[{Zhang and Li(2010)}]{zhang2010}
\bibinfo{author}{Zhang, Q.}, \bibinfo{author}{Li, B.}, \bibinfo{year}{2010}.
\newblock \bibinfo{title}{Discriminative k-svd for dictionary learning in face
  recognition}, in: \bibinfo{booktitle}{2010 IEEE Computer Society Conference
  on Computer Vision and Pattern Recognition}, pp. \bibinfo{pages}{2691--2698}.
\newblock \DOIprefix\doi{10.1109/CVPR.2010.5539989}.
%Type = Article
\bibitem[{Zheng et~al.(2001)Zheng, Zhang and Song}]{zheng2001comparison}
\bibinfo{author}{Zheng, F.}, \bibinfo{author}{Zhang, G.},
  \bibinfo{author}{Song, Z.}, \bibinfo{year}{2001}.
\newblock \bibinfo{title}{Comparison of different implementations of mfcc}.
\newblock \bibinfo{journal}{Journal of Computer Science and Technology}
  \bibinfo{volume}{16}, \bibinfo{pages}{582--589}.
\newblock \DOIprefix\doi{10.1007/BF02943243}.

\end{thebibliography}

\end{document}